%% file: main.tex
\documentclass[twoside]{article}

\usepackage[accepted]{aistats2021}

\makeatletter
\newcommand*{\addFileDependency}[1]{
  \typeout{(#1)}
  \@addtofilelist{#1}
  \IfFileExists{#1}{}{\typeout{No file #1.}}
}
\makeatother

\usepackage{amsfonts}
\usepackage{nicefrac}
\usepackage{amsmath}
\usepackage{amssymb}
\usepackage{amsthm}
\usepackage{amstext}
\usepackage{mathtools}
\usepackage{acronym}
\usepackage{bm}
\usepackage{url}
\usepackage{graphicx}
\usepackage[inline,shortlabels]{enumitem}
\usepackage{wrapfig}
\usepackage{pdfpages}
\usepackage{todonotes}
\usepackage{booktabs}
\usepackage[english]{babel}
\usepackage{caption}
\usepackage[export]{adjustbox}
\usepackage{subcaption}
\usepackage{mathptmx}  
\usepackage{natbib}
\usepackage{xr-hyper}
\usepackage{hyperref}
\usepackage{cleveref}

\usepackage{breqn}
\usepackage{subfiles}

\newtheorem{theorem}{Theorem}

\newtheorem{lemma}[theorem]{Lemma}

\newtheorem{innercustomgeneric}{\customgenericname}
\providecommand{\customgenericname}{}
\newcommand{\newcustomtheorem}[2]{%
  \newenvironment{#1}[1]
  {%
   \renewcommand\customgenericname{#2}%
   \renewcommand\theinnercustomgeneric{##1}%
   \innercustomgeneric
  }
  {\endinnercustomgeneric}
}

\newcustomtheorem{customthm}{Theorem}
\newcustomtheorem{customcorl}{Corollary}
\newcustomtheorem{customfct}{Fact}
\newcustomtheorem{customprop}{Proposition}

\acrodef{KLIEP}{\acl{KL} Importance Estimation Procedure}
\acrodef{CKLIEP}{Continual KLIEP}
\acrodef{LSIF}{Least-Squares Importance Fitting}
\acrodef{KL}{Kullback-Leibler}
\acrodef{EWC}{Elastic Weight Consolidation}
\acrodef{FID}{Fr{\'e}chet Inception Distance}
\acrodef{IS}{Inception Score}
\acrodef{KID}{Kernel Inception Distance}
\acrodef{VAE}{Variational Auto Encoder}
\acrodef{CVAE}{Continual \acl{VAE}}
\acrodef{CDRE}{Continual  \acl{DRE}}
\acrodef{TRE}{Telescoping   \acl{DRE}}
\acrodef{DRE}{Density Ratio Estimation}
\acrodef{MAE}{Mean Absolute Error}
\acrodef{GAN}{Generative Adversarial Network}
\acrodef{WGAN}{Wasserstein \ac{GAN}}
\acrodef{WGAN-GP}{\ac{WGAN} with Gradient Penalty}
\acrodef{VAE}{Variational Auto Encoder}
\acrodef{DVAE}{Discriminant VAE}
\acrodef{MMD}{Maximum Mean Discrepancy}
\acrodef{PRD}{Precision and Recall for Distributions}
\acrodef{VCL}{Variational Continual Learning}
\acrodef{IPM}{Integral Probability Metrics}
\acrodef{CNN}{Convolutional Neural Network}
\acrodef{PR}{Precision Recall}
\acrodef{VI}{Variational Inference}
\acrodef{MAE}{Mean Absolute Error}

\newcommand{\fgan}{{$f$-GAN}}
\newcommand{\fdiv}{{$f$-divergences}}
\newcommand{\cl}{{{continual learning}}}
\renewcommand{\sl}{{\emph{static learning}}}

\Crefname{equation}{Eq.}{Eqs.}
\Crefname{figure}{Fig.}{Figs.}
\Crefname{section}{Sec.}{Secs.}
\Crefname{algorithm}{Alg.}{Algs.}
\Crefname{theorem}{Thm.}{Thms.}
\Crefname{appendix}{Appx.}{Appxs.}
\Crefname{table}{Tab.}{Tabs.}

\begin{document}

\twocolumn[

\aistatstitle{\acl{CDRE} in an Online Setting}

\aistatsauthor{ Yu Chen \And Song Liu \And  Tom Diethe \And Peter Flach }

\aistatsaddress{ University of Bristol \And  University of Bristol \And Amazon Research \And University of Bristol} ]

\begin{abstract}
In online applications with streaming data, 
awareness of how far the training or test set has shifted away from the original dataset can be crucial to the performance of the model. However, we may not have access to historical samples in the data stream. To cope with such situations,  
we propose a novel method,  \ac{CDRE}, for estimating density ratios between the initial and current distributions ($p/q_t$) of a data stream in an iterative fashion without the need of storing past samples, where $q_t$ is shifting away from $p$ over time $t$. We demonstrate that \ac{CDRE} can be more accurate than standard \ac{DRE} in terms of estimating divergences between distributions, despite not requiring samples from the original distribution. \ac{CDRE} can be applied in scenarios of online learning, such as importance weighted covariate shift, tracing dataset changes for better decision making.  
In addition, \ac{CDRE} enables the evaluation of generative models under the setting of \cl{}. To the best of our knowledge, there is no existing method that can evaluate generative models in  continual learning without storing samples from the original distribution.
 
\end{abstract}


\input{./sec1}

\input{./sec2}

\input{./sec3}

\input{./sec4}

\clearpage
\bibliographystyle{apalike}
\bibliography{references}
\clearpage
\input{supp}

\end{document}

%% file: sec1.tex
\section{Introduction}

In the real world, online applications are ubiquitous in practice since large amounts of data are generated and processed in a streaming manner. There are two types of machine learning scenarios commonly deployed for such streaming data:
\begin{enumerate}[label={\arabic*})]
    \item train a model online on the streaming data (\emph{e.g.} online learning \citep{shalev2012online} and continual learning \citep{parisi2019continual}) -- in this case the training set may be shifting over time;
    \item train a model offline and deploy it online -- in this case the test set may be shifting over time.
\end{enumerate}
In both cases, the main problem  is dataset shifting, i.e. the data distribution changes gradually over time. Awareness of how far the training or test set has been shifted can be crucial to the performance of the model.
For example, when the training set is shifting, the latest model may become less accurate on samples from earlier data distributions, like the covariate shift \citep{shimodaira2000improving}. In this case, we can use importance weights to `rollback' the model for a better prediction, where the importance weights are  density ratios. In the other case, the performance of a pre-trained model may gradually degrade when the test set shifts away from the training set over time. It will be beneficial to trace the distribution difference caused by dataset shifting so that we can decide when to update the model for preventing the performance from degrading.

\acf{DRE} \citep{sugiyama2012density} is a method for estimating the ratio between two probability distributions which can reflect the difference between the  two distributions.  In particular, it can be applied to settings in which only samples of the two distributions are available, which is usually the case in practice. However, under certain restrictive conditions in online applications -- e.g., unavailability of historical samples in an online data stream -- existing \ac{DRE} methods are no longer applicable. Moreover, \ac{DRE} exhibits difficulties for  accurate estimations when there exists significant differences between the two distributions  \citep{sugiyama2012density, mcallester2020formal,rhodes2020telescoping}. 
In this paper, we propose a new framework of density ratio estimation called \acf{CDRE} 
which is capable of coping with the online scenarios and gives  better estimation than standard \ac{DRE} when the two distributions are less similar. There are existing methods for detecting changing points online by \ac{DRE} \citep{kawahara2009change,liu2013change,bouchikhi2018non}, which estimate density ratios between distributions of  two consecutive time intervals. In contrast, \ac{CDRE} estimates density ratios between distributions of the initial and latest time intervals without storing historical samples. 

\ac{CDRE} can be applied to tracing the differences between distributions by estimating their \fdiv{} and thus it provides a new option for evaluating generative models in continual learning.
The scenario of the training set shifting over time matches the problem setting of \cl{} \citep{parisi2019continual} in which a single model is trained by a set of tasks sequentially with no (or very limited) access to the data from past tasks, and yet is able to perform on all learned tasks. 
The existing measures of  evaluating generative models are basically to estimate the
difference between the original data distribution and the distribution of model samples  \citep{heusel2017gans,binkowski2018demystifying}.  
All those methods require the samples from the original data distribution which may not be possible in the setting of \cl{}, 
yet \ac{CDRE} can fit in such a situation. 

Our key contributions in this paper are: 
\begin{enumerate}[label=\emph{\roman*})]
    \item we propose a new framework \ac{CDRE} for estimating density ratios in an online setting, which does not require storing historical samples in the data stream; 
    
    \item we provide an instantiation of \ac{CDRE} by using \ac{KLIEP} \citep{sugiyama2008direct} as a building block, and provide theoretical analysis of its asymptotic behaviour;
    
    \item we demonstrate the efficacy of \ac{CDRE} in several online  applications, including backward covariate shift,  tracing distribution drift, and evaluating generative models in \cl{}. 
    To the best of our knowledge, there is no prior work that can evaluate generative models in continual learning without storing samples from the original distribution.
\end{enumerate}

The rest of this paper is structured as follows. 
\Cref{sec:kliep} briefly reviews the formulation of  \ac{KLIEP}, and \Cref{sec:ol_setting} introduces the problem setting of \ac{CDRE}.
In \Cref{sec:CDRE} we provide the technique details of \ac{CDRE} and demonstrate the instantiation of \ac{CDRE} by \ac{KLIEP}.   
\Cref{sec:experiment} introduces several applications of \ac{CDRE} along with comprehensive  experimental results. 
Finally, \Cref{sec:discuss} provides some further discussion about \ac{CDRE} and its applications.

%% file: sec2.tex
\section{Preliminaries}\label{sec:background}

We first review the formulation of \ac{KLIEP} which we use for instantiating \ac{CDRE}. We then formally introduce the problem setting of \ac{CDRE}. 

\subsection{\acl{DRE} by \acl{KLIEP}}\label{sec:kliep}

\ac{KLIEP} is a classic method for density ratio estimation  introduced in \cite{sugiyama2008direct}.    
Here we review the formulation of \ac{KLIEP} because we deploy it as an example of the basic estimator of \ac{CDRE}. 

Let $r^*(x) = \frac{p(x)}{q(x)}$ be the (unknown) true density ratio, then $p(x)$ can be estimated by $\tilde{p}(x) = r(x)q(x)$, where $r^*(x)$ is modeled by $r(x)$. Hence, we can optimize $r(x)$ by minimizing the 
\ac{KL}-divergence between $p(x)$ and $\tilde{p}(x)$ with respect to $r$:
\begin{equation}\label{eq:kliep}
    \begin{split}
        & D_{KL}\left(p(x)||\tilde{p}(x)\right) = \int p(x) \log \frac{p(x)}{\tilde{p}(x)} dx \\
        & \quad = \int p(x) \log r^*(x) dx - \int p(x) \log r(x) dx
    \end{split}
\end{equation}
where $r(x)$ should satisfy $r(x) > 0$ and $\int r(x)q(x) dx = \int \tilde{p}(x) dx = 1$.
As the first term of the right-hand side in \Cref{eq:kliep} is constant w.r.t. $r(x)$, the empirical objective of optimizing $r(x)$ is as follows:
\begin{equation}\label{eq:kliep_obj}
    \begin{split}
        J_r &= \max_r \frac{1}{N}\sum_{i=1}^N \log r(x_i), \ \ x_i \sim p(x), \\
         & s.t. \ \ \frac{1}{M}\sum_{j=1}^M r(x_j) = 1, \ \  
         r(x) \ge 0, \ \ x_j \sim q(x). 
    \end{split}
\end{equation}
One convenient way of parameterizing $r(x)$ is by using a log-linear model with normalization, which then automatically satisfies the constraints in \Cref{eq:kliep_obj}:
\begin{equation}\label{eq:loglinear_r}
    \begin{split}
        & r(x;\beta) = \frac{\exp(\psi_{\beta}(x))}{\frac{1}{M} \sum_{j=1}^M \exp(\psi_{\beta}(x_j))} ,  \\
        & x_j \sim q(x), \quad
        \psi_{\beta}: \mathbb{R}^D \rightarrow \mathbb{R},
    \end{split}
\end{equation}
where $\psi_{\beta}$ can be any deterministic function: we use a neural network as $\psi_{\beta}$ in our implementations, $\beta$ then  representing parameters of the neural network.

\subsection{The problem setting of \ac{CDRE}}\label{sec:ol_setting}

The goal of \ac{CDRE} is  estimating density ratios between two distributions $r_{\tau,t}(x) = p_{\tau}(x)/q_{\tau,t}(x)$, where $\tau$ is the initial time index of starting tracing a distribution, $t$ denotes the current time index, and when $t > \tau$ the samples of $p_{\tau}(x)$ is unavailable. We refer to $p_{\tau}(x)$ as the \emph{original distribution}  and $q_{\tau,t}(x)$ as the \emph{dynamic distribution} of $p_{\tau}(x)$. The dynamic distribution is assumed to be shifting away from its original distribution gradually over time. For example, let $\tau = 1$, at $t = 1$ we have access to samples of both $p_\tau$ and $q_{\tau,t}$, we can directly estimate $r_{\tau,t}$ by standard \ac{DRE}, and when $t>1$, we have no access to samples of $p_\tau$ any longer, instead, we only have access to samples of $q_{\tau,t-1}$ and $q_{\tau,t}$. \ac{CDRE} is proposed to estimate $r_{\tau,t}$ in such a situation. 
In addition, \ac{CDRE} is able  to  estimate ratios of multiple pairs of the original and dynamic distributions by a single estimator. It avoids building separated estimators for tracing different original distributions in an application (e.g. seasonal data), which also fits the common setting of \cl{} that we will introduce in the latter sections.

%% file: sec3.tex


\section{\acl{CDRE}}\label{sec:CDRE}

In this section we introduce the basic formulation of \ac{CDRE} and demonstrate instantiating it by \ac{KLIEP}. For convenience, we initially introduce the formulation with a single pair of original and dynamic distributions. We then give a more general  formulation for multiple pairs of original and dynamic  distributions. 

\subsection{The basic form of \ac{CDRE}}\label{sec:CDRE_single}

For simplicity of notation, we assume $\tau = 1$ in the case of estimating density ratios between a \emph{single} pair of original and dynamic distributions, and then omit $\tau$ in the basic  formulations. Thus, the density function of the original distribution is denoted as $p(x)$  and  its samples are  unavailable when $t > 1$. Similarly,  $q_{t}(x)$ denotes the density function of the dynamic distribution at time $t$. The true
density ratio $r^*_t(x) \triangleq p(x)/q_t(x)$ can be decomposed as follows:
\begin{equation}\label{eq:continual_ratio}
    \begin{split}
        & r^*_t(x)
        = \frac{q_{t-1}(x)}{q_{t}(x)}
        \frac{p(x)}{q_{t-1}(x)}
        = r^*_{s_{t}}(x) r^*_{t-1}(x),\ \ t > 1,
    \end{split}
\end{equation}
where $r^*_{s_{t}}(x) \triangleq  q_{t-1}(x)/q_{t}(x)$ represents the true density ratio between the 
two latest dynamic distributions. Using this decomposition we can 
estimate $p(x)/q_t(x)$ in an  iterative manner without the need of storing samples from $p(x)$ when $t$ increases.
The key point is that we can estimate $r^*_t(x)$ by estimating  $r^*_{s_t}(x)$ when the estimation of $r^*_{t-1}(x)$ is known. In particular, it introduces one extra constraint: 
\begin{equation}\label{eq:cl_constr}
	\begin{split}
		\int r^*_{s_t}(x) q_t(x) dx = \int \frac{r^*_{t}(x)}{r^*_{t-1}(x)} q_t(x) dx = 1 
	\end{split}
\end{equation} 
Existing methods of \ac{DRE} can be applied to estimating the initial ratio $r^*_1(x)=p(x)/q_1(x)$ and the latest ratio $r^*_{s_t}(x), \forall t>1$, as the basic ratio estimator of \ac{CDRE}. Let $r_t(x)$ be the estimation of $r^*_t(x)$, where $r_{t-1}$ is already obtained, then the objective of \ac{CDRE} can be expressed as: 
\begin{equation}\label{eq:fd-form}
    \begin{split}
        & J_{CDRE}(r_t) = J_{DRE}(r_{s_t}), \ \ 
        r_{s_t}(x) \triangleq \frac{r_{t}(x)}{r_{t-1}(x)} \\
        &
        s.t. \frac{1}{N}\sum_{n=1}^N r_{s_t}(x_n) = 1, \ \ 
        x_n \sim q_t(x).
    \end{split}
\end{equation}
where $J_{DRE}$ can be the objective of any method used for standard \ac{DRE}, such as \ac{KLIEP} \citep{sugiyama2008direct}.

\subsection{An instantiation of \ac{CDRE}: CKLIEP}
We now demonstrate how to instantiate \ac{CDRE} by \ac{KLIEP}, which we call \ac{CKLIEP}. Define $r_t(x), r_{t-1}(x)$ by  the log-linear form as in \Cref{eq:loglinear_r}, let $N_t = N_{t-1} = N$ as the sample size of each distribution, then $r_{s_{t}}$ is as follows:
\begin{equation}\label{eq:cur_ratio}
    \begin{split}
        & r_{s_{t}} 
        =  {\exp\{\psi_{\beta_t} (x) - \psi_{\beta_{t-1}} (x) \}}
       \times \frac{\frac{1}{N}\sum_{j=1}^{N} \exp\{\psi_{\beta_{t-1}}(x_{t-1,j})\}}
        {\frac{1}{N}\sum_{i=1}^{N} \exp\{\psi_{\beta_t}(x_{t,i})\}} , \\
        & \quad x_{t,i} \sim q_{_t}(x), \quad x_{t-1,j} \sim q_{{t-1}}(x).
    \end{split}
\end{equation}
where $\beta_t,\beta_{t-1}$ represent parameters of
$r_t(x), r_{t-1}(x)$, respectively. 
When the constraint in \Cref{eq:fd-form} is satisfied, we have the following equality by substituting \Cref{eq:cur_ratio} into the constraint:
\begin{equation}\label{eq:cl_logl_constr}
    \begin{split}
        &\frac{{ \sum_{i=1}^{N} \exp\{\psi_{\beta_t}(x_{t,i})\}}}{ \sum_{j=1}^{N} \exp\{\psi_{\beta_{t-1}}(x_{t-1,j})\}} 
         = \frac{1}{N}\sum_{i=1}^{N} \exp\{\psi_{\beta_t}(x_{t,i}) - \psi_{\beta_{t-1}}(x_{t,i})\}
    \end{split}
\end{equation}
$r_{s_t}$ can then be rewritten in the same log-linear form of \Cref{eq:loglinear_r} by substituting  \Cref{eq:cl_logl_constr} into \Cref{eq:cur_ratio}:
\begin{equation}\label{eq:cur_ratio2}
    \begin{split}
        r_{s_{t}}
        &=  \frac{\exp\{\psi_{\beta_t} (x) - \psi_{\beta_{t-1}} (x) \}}
        {\frac{1}{N}\sum_{i=1}^{N} \exp\{\psi_{\beta_t}(x_{t,i}) - \psi_{\beta_{t-1}}(x_{t,i})\}}
        \\
        &= \frac{\exp\{\phi_{\beta_t}(x)\}}{\frac{1}{N}\sum_{i=1}^{N}\exp\{\phi_{\beta_t}(x_i)\}}, \\
        &\text{where} \ \ \phi_{\beta_t}(x) \triangleq \psi_{\beta_t}(x) - \psi_{\beta_{t-1}}(x).
    \end{split}
\end{equation}
Now we can instantiate $J_{DRE}$ in \Cref{eq:fd-form} by the objective of \ac{KLIEP} (\Cref{eq:kliep_obj}) and adding the equality constraint (\Cref{eq:cl_logl_constr}) into the objective with a hyperparameter $\lambda_{c}$, which gives the objective of \ac{CKLIEP} as the following:
\begin{equation}\label{eq:loss_ck}
\begin{split}
    & \max_{\beta_t}\mathcal{L}_t(\beta_t) = \max_{\beta_t}  \frac{1}{N} \sum_{j=1}^{N} \log r_{s_t}(x_{t-1,j}) \\
    & \qquad \qquad \qquad \qquad +  \lambda_{c}
    \left(\frac{\Psi_t(x_t)}{\Phi_t(x_t)\Psi_{t-1}(x_{t-1})}-1\right)^2\\
    & \text{where} \ \ t > 1,\ \ x_{t,i} \sim q_t(x),\ \  x_{t-1,j} \sim q_{t-1}(x),
     \\
     &
     \Phi_{t}(x_t) \triangleq  \frac{1}{N}\sum_{i=1}^{N} \exp\{\phi_{\beta_t}(x_{t,i})\},  
    \Psi_t(x_t) \triangleq {\frac{1}{N} \sum_{i=1}^{N} \exp\{\psi_{\beta_t}(x_{t,i})\}}
\end{split}
\end{equation}
where $\beta_{t-1}$ is the estimated parameter of $r_{t-1}(x)$ and hence a constant in the objective.

A concurrent work \citep{rhodes2020telescoping} has developed \ac{TRE} by the same  consecutive decomposition in \Cref{eq:continual_ratio}  but with the following main differences: 
\begin{enumerate*}[label=\emph{\arabic*})]
    \item the objective of \ac{TRE} is to optimize a set of ratio estimators at the same time whereas \ac{CDRE} only needs to  optimize the latest ratio  estimator; 
    \item \ac{TRE} requires samples of all intermediate  distributions as well as the original distribution,  in contrast \ac{CDRE} only requires samples of the two latest distributions. 
\end{enumerate*}


\input{theory}


\subsection{Multiple original distributions in \ac{CDRE}}

Now we consider tracing  multiple original distributions in \ac{CDRE}, in which case a new pair of original and dynamic distributions will be added into the training process of the estimator at some time point. We refer to an original distribution as $p_{\tau}(x)$, where $\tau$ is the time index of starting tracing the original distribution. And  samples of $p_{\tau}(x)$ are not available when $t > \tau$. Similarly,  $q_{\tau,t}(x)$ denotes the 
dynamic distribution that  corresponding to $p_{\tau}(x)$ at time $t$:
\begin{equation*}\label{eq:continual_ratio_full}
    \begin{split}
         r^*_{\tau,t}(x) &= \frac{p_{\tau}(x)}{q_{\tau,t}(x)} 
        = \frac{q_{\tau,t-1}(x)}{q_{\tau,t}(x)}
        \frac{p_{\tau}(x)}{q_{\tau,t-1}(x)} 
        = r^*_{s_{\tau,t}}(x) r^*_{\tau,t-1}(x),
    \end{split}
\end{equation*}
Where $r^*_{s_{\tau,t}}(x) = { q_{\tau,t-1}(x)}/{q_{\tau,t}(x)}$. In this case, we optimize the estimator at time $t$ by an averaged objective:
\begin{equation}
    \begin{split}
        \max_{\beta_t}  \bar{\mathcal{L}}_t(\beta_t) = \max_{\beta_t} \frac{1}{|\mathbb{T}|} \sum_{\tau \in \mathbb{T}} \mathcal{L}_t(\beta_t;\tau)
    \end{split}
\end{equation}
where $\mathbb{T}$ is the set of time indices of adding original distributions, $|\mathbb{T}|$ is the size of $\mathbb{T}$.  $\mathcal{L}_t(\beta_t;\tau)$ is as the same as the loss function of a single original distribution (  \Cref{eq:loss_ck}) for a given $\tau$. Further,  $r_{s_{\tau,t}}(x)$ is also  defined by the same form of \Cref{eq:cur_ratio2}, the difference is that  $\psi_{\beta_t}(x)$ becomes  $\psi_{\beta_t}(x;\tau)$. In our implementation, we  concatenate the time index $\tau$ to each data sample as the input of the ratio estimator. Thus,  we can avoid learning separate ratio estimators for multiple original distributions. 
In addition, we set the output of $\psi_{\beta_t}(\cdot)$  as a $|\mathbb{T}|$-dimensional vector $\{o_1,\dots,o_i,\dots,o_{|\mathbb{T}|} \}$ where $o_{i}$ corresponds to the output of $\psi_{\beta_t}(x;\tau=\mathbb{T}_i)$. 
Note that with \ac{CDRE} we have the flexibility to extend the model architecture since the latest estimator function $\phi_{\beta_t}$ only needs the output of the previous estimator function $\psi_{\beta_{t-1}}$. This can be beneficial when the model capacity becomes a bottleneck of the performance.





\subsection{Dimensionality  reduction in online applications}

The main concern of estimating  density ratios stems from high-dimensional data. Several methods for dimensionality reduction in \ac{DRE} have been introduced in \citep{sugiyama2012density}. A fundamental assumption of these methods is that the difference between two distributions can be confined to a subspace, which means ${p(z)}/{q(z)} = {p(x)}/{q(x)}$ where $z$ is a lower-dimensional representation of $x$. This aims for exact density ratio estimation in a subspace but incurs a high computational cost. 
In most applications, a model (e.g. classifiers) is often trained upon a representation extractor, indicating a surrogate feature space of high-dimensional data can be used for the density ratio estimation in practice. For instance, most measurements of generative models estimate the difference between two distributions in a surrogate feature space, e.g., the inception feature defined for the \ac{IS} \citep{salimans2016improved} is extracted by a neural network, and this method is also applied in many other measurements of generative models, e.g. \ac{FID} \citep{heusel2017gans}, \ac{KID} \citep{binkowski2018demystifying}.

In prior work, a pre-trained classifier is often used to generate surrogate features of high-dimensional image data (e.g. inception features \citep{salimans2016improved}). However, it may be difficult to train such a classifier in online applications because: a) a homogeneous dataset for all unseen data or tasks may not be available in advance; b) labeled data may not be available in generative tasks. In order to cope with such circumstances, we
introduce \acf{CVAE} in a pipeline with \ac{CDRE}. 
The loss function of \ac{CVAE} is defined as the following:
\begin{equation}
    \begin{split}
        &\mathcal{L}_{CAVE}(\theta_t,\vartheta_t) = NLL
        + \mathcal{D}_{KL}(q_t(z)||q_{t-1}(z)),\\
        &NLL = - \frac{1}{t}\biggl[\sum_{\tau=1}^{t-1} \mathbb{E}_{q_{\tau,t-1}(x)}[\mathbb{E}_{q_t(z)}[\log p(x|z;\vartheta_t)]]  \\
        & \qquad \qquad + \mathbb{E}_{p_t(x)}[\mathbb{E}_{q_t(z)}[\log p(x|z;\vartheta_t)]]\biggr]
    \end{split}
\end{equation}
where $q_t(z) = \mathcal{N}(\mu_{\theta_t}(x),\sigma_{\theta_t}(x))$, $\theta_t$ and $\vartheta_t$ denote parameters of the encoder and decoder of \ac{CVAE}, respectively. $NLL$ is the negative log likelihood term as the same as in vanilla \acs{VAE} \citep{kingma2013auto}, the training data set includes samples of all dynamic distributions at the previous step ($q_{\tau,t-1}(x), \forall \tau < t$) and the  samples of the current original distribution  ($p_t(x)$). 
The \ac{KL}-divergence term in the loss function serves a regularization term: the current encoder is expected to give similar $z$ for a similar $x$ comparing with the previous encoder. This term forces the consistency between inputs of the previous and current ratio estimators (i.e. inputs of  $\psi_{\beta_{t-1}}$ and $\psi_{\beta_{t}}$ in \Cref{eq:loss_ck}). \ac{CVAE} can generate effective features without requiring labels or pre-training, nevertheless, other commonly used methods (e.g. pre-trained classifiers) are also applicable to \ac{CDRE}.

%% file: theory.tex
\subsection{Asymptotic normality of \ac{CKLIEP}}
Define $\hat{\beta}_t$ as the estimated parameter that satisfies: 
\begin{equation}\label{eq:def_hat}
    \mathcal{L}^'_t(\hat{\beta}_t) \triangleq \nabla_{\beta_t} \mathcal{L}_t(\beta_t)\big|_{\beta_t = \hat{\beta}_t} =   0
\end{equation}

 Assume $\phi_{\beta_t}(x)$ (\Cref{eq:cur_ratio2}) includes the correct function that there exists $\beta_{t}^*$ recovers the true ratio over the  population:
\begin{equation}\label{eq:def_optm}
    \begin{split}
     & r^*_{s_t}(x)  
    = \frac{q_{t-1}(x)}{q_t(x)}
    = \frac{\exp\{\phi_{\beta^*_t}(x)\}}{\mathbb{E}_{ q_t}[\exp\{\phi_{\beta^*_t}(x)\}]}, \\
    & \text{where} \ \ \phi_{\beta^*_t}(x) = \psi_{\beta_t^*}(x) - \psi_{{\beta}_{t-1}}(x),
    \end{split}
\end{equation}

\textbf{\emph{Notations}}: $\leadsto$ and $\xrightarrow{P}$ mean convergence in distribution and convergence in probability, respectively. 

\textbf{\emph{Assumptions}}: We assume $q_t(x)$ and $q_{t-1}(x)$ are independent, $n_t = n_{t-1} = n$, where $n_t$ is the sample size of $q_t(x)$. Let $S_t$ be the support of $q_t$, we assume $S_{t-1} \subseteq S_t$ in all cases. 

\begin{lemma}\label{lemma1}
  Let $\ell^'_r(\beta^*_t) \triangleq \frac{1}{n}\sum_{j=1}^{n}\nabla_{\beta_t} \log r_{s_t}(x_{t-1,j})|_{\beta_t = \beta^*_t}$, we have   $\sqrt{n}\ell^'_r(\beta^*_t) \leadsto \mathcal{N}(0, \sigma^2)$, where   
\begin{equation*}
\begin{split}
 \sigma^2 &= Cov_{q_{t-1}}[\nabla_{\beta_t}\phi_{\beta^*_t}(x)] + 
 \frac{Cov_{q_t}[\nabla_{\beta_t}\exp\{\phi_{\beta^*_t}(x)\}]}{\mathbb{E}_{q_t}[\exp\{\phi_{\beta^*_t}(x)\}]^2}
\end{split}
\end{equation*}
\end{lemma}

\begin{lemma}\label{lemma2}
Let $\ell^{''}_r(\beta^*_t) \triangleq \frac{1}{n} \sum_{j=1}^{n} \nabla^2_{\beta_t}\log r_{s_t}(x_{t-1,j})|_{\beta_t=\beta^*_t}$, we have  $\ell^{''}_r(\beta^*_t) \xrightarrow{P} -I_{\beta^*_t}$, where ${I}_{\beta^*_t} \triangleq Cov_{q_{t-1}}
[\nabla_{\beta_t}\phi_{\beta^*_t}(x)]$.
\end{lemma}

\begin{lemma}\label{lemma3}
Let $\ell_c(\beta_t) \triangleq \lambda_c \left(\frac{\Psi_t(x_t)}{\Phi_t(x_t)\Psi_{t-1}(x_{t-1})}-1\right)^2$, and $ \ell^'_c(\beta^*_t) \triangleq  \nabla_{\beta_t}\ell_c(\beta_t)|_{\beta_t = \beta^*_t} $, if we set $\lambda_c = \frac{A}{\sqrt{n}}$, where $A$ is a positive constant, then $\sqrt{n}\ell^'_c(\beta^*_t) \xrightarrow{P} 0$.
\end{lemma}

\begin{customthm}{1}\label{thm1}
Suppose $\lambda_c = \frac{A}{\sqrt{n}}$, where $A$ is a positive constant, assume  $\ell^{''}_c(\beta^*_t) = o_p(1)$, $\hat{\beta_t}-\beta^*_t=o_p(1)$, $\mathcal{L}^{'''}_t(\tilde{\beta}_t) = O_p(1)$, where $\tilde{\beta}_t$ is a point between $\hat{\beta}_t$ and $\beta^*_t$, then $\sqrt{n} (\hat{\beta_t} - \beta_t^*) \leadsto \mathcal{N}(0, \nu^2)$, 
where 
\begin{equation}
    \begin{split}
      \nu^2 &=  I^{-1}_{\beta^*_t} + \mathbb{E}_{q_t}[\exp\{\phi_{\beta^*_t}(x)\}]^{-2} \times \\
     & \quad 
     I^{-1}_{\beta^*_t} Cov_{q_t}[\nabla_{\beta_t}\exp\{\phi_{\beta^*_t}(x)\}]
     I^{-1}_{\beta^*_t}
    \end{split}
\end{equation}
\end{customthm}

\begin{figure}
    \centering
    \includegraphics[width=0.65\linewidth,trim={0.2cm 0cm 1.5cm 1.cm},clip]{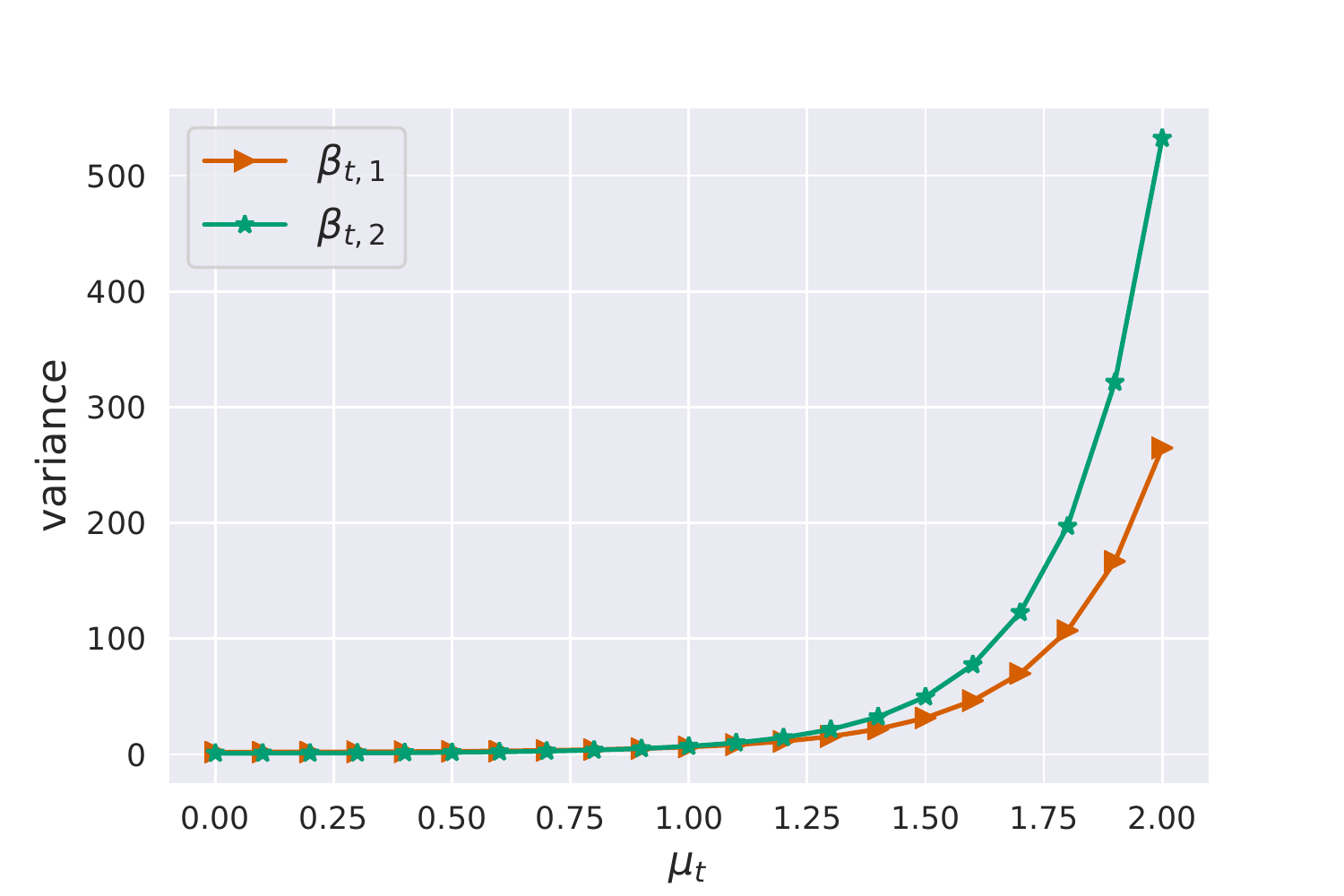}
    \caption{Demonstration of the variance of estimated parameters in Corollary 1  by 1-D Gaussian distributions: fix  $q_{t-1}(x) = \mathcal{N}(0,1)$ and  $q_t(x) = \mathcal{N}(\mu_t, 1)$.}
    \label{fig:corl1}
\end{figure}

\begin{customcorl}{1}\label{corl1}
Suppose $p(x)$ and $\forall t, q_t(x)$ are from the exponential family, define $r^*_{s_t}(x) = \exp\{\phi_{\beta^*_t}(x)\}$,  $\phi_{\beta^*_t}(x) = \beta^*_{t} T(x)+C$,  $T(x)$ is a sufficient statistic of $x$, $C$ is a constant, then  $\sqrt{n} (\hat{\beta_t} - \beta_t^*) \leadsto \mathcal{N}(0, \nu_e^2)$, where $T(x)$ is a column vector, $T(x)^2 = T(x)T(x)^T$, $I_{\beta^*_t}=Cov_{q_{t-1}}[{T}(x)]$:
\begin{equation}\label{eq:corl1}
    \begin{split}
        \nu_e^2 &= I^{-1}_{\beta^*_t} + I^{-1}_{\beta^*_t}  ({\mathbb{E}_{q_{t-1}}[r^*_{s_t}(x){T}(x)^2]-\mathbb{E}_{q_{t-1}}[{T}(x)]^2})I^{-1}_{\beta^*_t}
    \end{split}
\end{equation}
\end{customcorl}
All proofs are provided in \Cref{sec:proof}.  Corollary 1 shows that how the covariance matrix $\nu_e^2$ depends on the latest density ratio ($r^*_{s_t}$) when the  distributions are from the exponential family. Since a smaller variance is better for convergence, we would prefer $r^*_{s_t}(x) = q_{t-1}(x)/q_t(x)$ is small, which means  when $q_{t-1}(x)$ is large $q_t(x)$ should be also large. In this case, $r^*_{s_t}$ is less likely to explode and the variance of the estimated parameter would be likely confined. We demonstrate this by experiments with 1-D Gaussian distributions. We fix $q_{t-1}(x) = \mathcal{N}(0,1)$, setting $q_t(x) = \mathcal{N}(\mu_t, 1)$, where $\mu_t=\delta k, \delta=0.1, k \in \{0,1,\dots,20\}$. In this case, $T(x) = \{x,x^2\}$, $\beta_t=\{\beta_{t,1},\beta_{t,2}\}$, we display the diagonal of  $\nu_e^2$ (variance of $\beta_{t,1}, \beta_{t,2}$) in \Cref{fig:corl1}. It is clear that when $q_t$ is farther to $q_{t-1}$, the variance  is larger. 

When $\beta_t = \beta^*_t, n\rightarrow \infty$, we have $r_t(x) = r^*_{s_t}(x)r_{t-1}(x)$, then $\log r^*_t(x) - \log r_t(x) = \log r^*_{t-1}(x) - \log r_{t-1}(x)$, which means the error inherited from $r_{t-1}(x)$ will be the intrinsic error for estimating $r^*_t(x)$ due to the objective of \ac{CDRE} is iterative. It indicates that smaller difference between each intermediate $q_\tau$ and $q_{\tau-1}$  ($\forall 1 < \tau \le t$) leads to a better estimation.  
We demonstrate in \Cref{sec:app_trace}  that it is often the case in practice even when $\psi_{\beta^*_t}(x)$ is approximated by a non-linear model. Moreover, \Cref{fig:corl1} shows the variance of the estimation may grow rapidly when the difference between the two distributions exceeds a certain value. \ac{CKLIEP} can prevent such an issue by ensuring the difference between any  intermediate pairs of distributions is relatively small. We will demonstrate this in \Cref{sec:app_trace} as well.  

%% file: sec4.tex
\section{Applications}\label{sec:experiment}
\begin{figure}[t!]
    \centering
    \begin{subfigure}{0.48\linewidth}
    \includegraphics[width=\linewidth,trim={0.5cm .2cm 1.5cm 1.cm},clip]{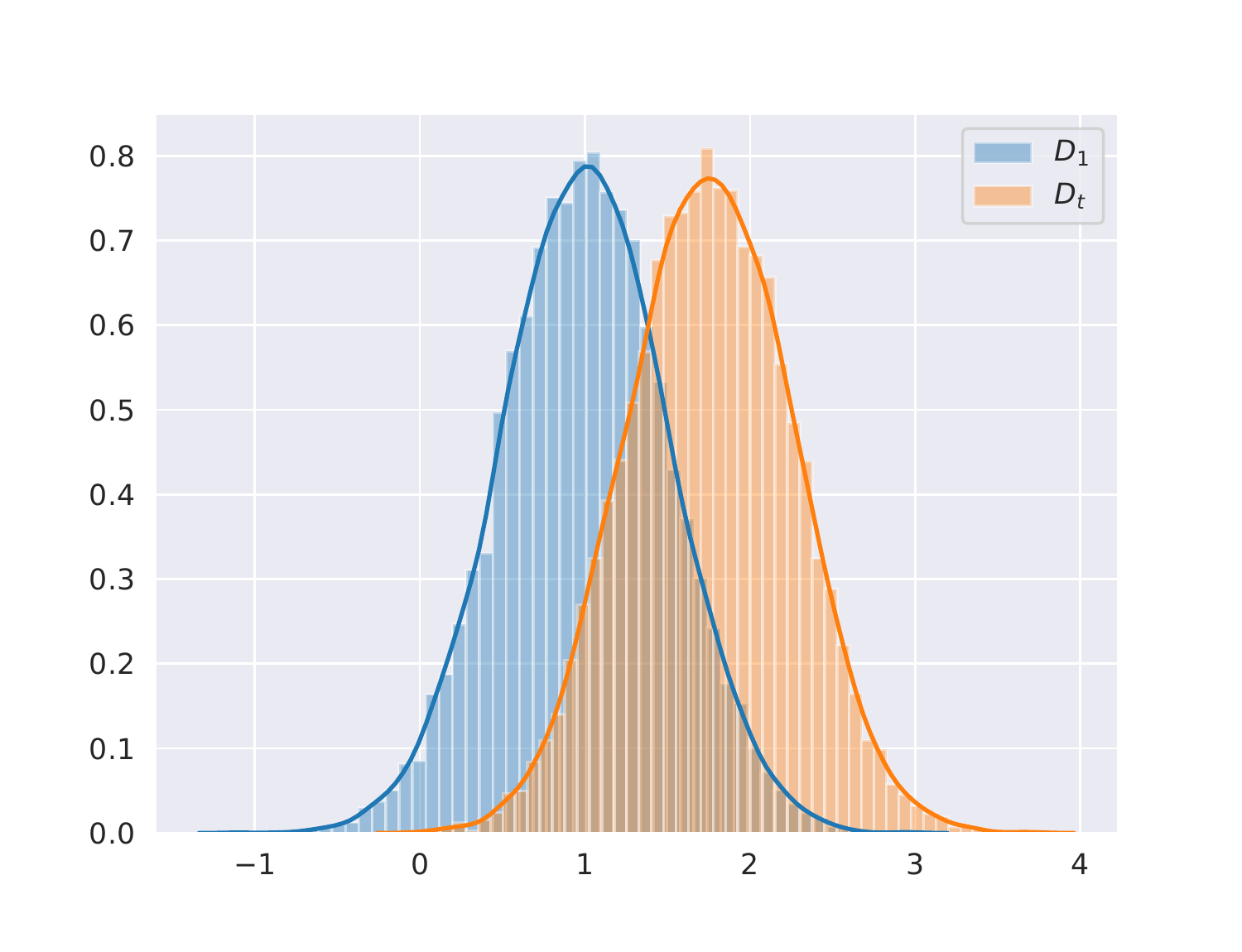}
    \caption{data distribution at $\tau=1$ ($D_1$) and $\tau=t$ ($D_t$)}
    \label{fig:reg_dist}
    \end{subfigure}
    \hfill%
    \begin{subfigure}{0.48\linewidth}
    \includegraphics[width=\linewidth,trim={0.8cm .2cm 1.5cm 1.cm},clip]{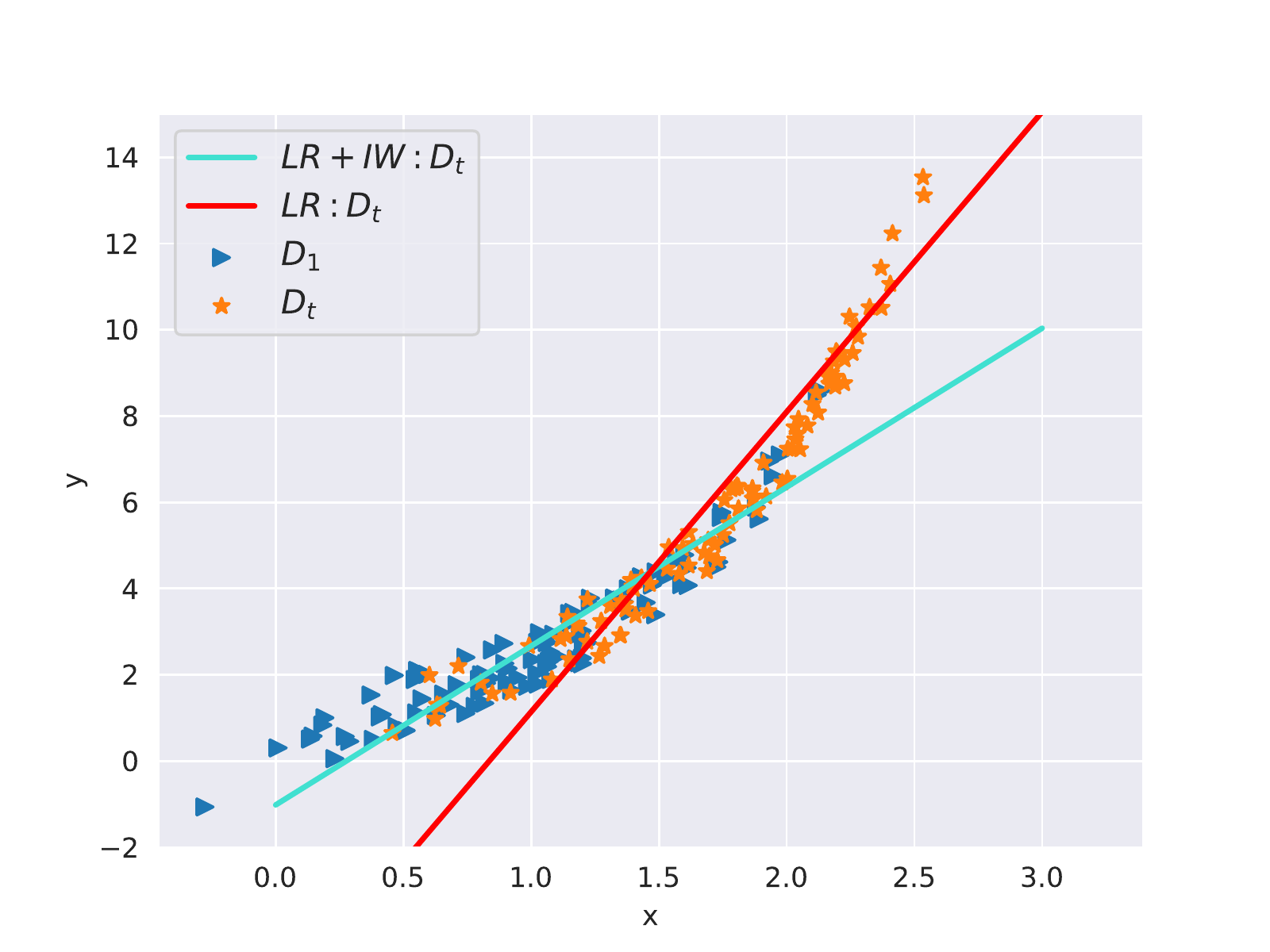}
    \caption{Linear regression trained by $D_t$ for fitting $D_1$ }
    \label{fig:reg_line}
    \end{subfigure}
\caption{Demo experiment of backward covariate shift. (a) shows the data distribution of training set at $\tau=1$ and $\tau=t$. (b) displays the regression lines learnt by the model at $\tau=t$, the cyan and red lines are fitted by $D_t$ with and without importance weights, respectively.}
\label{fig:toy_cl}
\end{figure}

In this section we introduce several applications of \ac{CDRE}:
\begin{enumerate*}[label={\arabic*})]
    \item backwards covariate shift;
    \item tracing distribution shifts by KL-divergence;
    \item evaluating generative models in continual learning.
\end{enumerate*} 
We demonstrate the efficacy of \ac{CDRE} by comprehensive experiments. The standard deviation of all results are from 10 runs with different random seeds. In all of our experiments,  $\psi(\cdot)$ is a neural network with two dense layers, each having 256 hidden units and ReLU activations. We provide more  details of the experiments in the supplementary material.

\subsection{Backwards covariate shift}

We demonstrate that \ac{CDRE} can be applied in \emph{backward  covariate shift} in which case the training set is shifting and the test set is from a previous distribution. It just swaps the situation of training and test set in common covariate shift \citep{sugiyama2008direct,shimodaira2000improving}. We assume a linear regression model defined as $\hat{y}=wx+b+\epsilon_0$, where the noise $\epsilon_0 \sim \mathcal{N}(0,0.01)$. At time $\tau$, the training data $x \in D_\tau$ and $D_\tau$ is shifting away from $D_1$ gradually, where $\tau \in \{1,2,\dots,t\}$ and $t=10$ is the latest time index. \Cref{fig:reg_dist} displays the data distribution at time $\tau = 1$ and $\tau=t$ in which we can see there exists notable difference between the two distributions. When the model is trained by $D_t$, it will not be able to accurately predict on test samples from $D_1$ unless we adjust the loss function by importance weights (i.e. density ratios) as in handling  covariate shift:
\begin{equation*}
    \mathcal{L} = \mathbb{E}_{x \sim q_t(x)}\left[\frac{q_1(x)}{q_t(x)}(y-\hat{y})^2\right]
\end{equation*}
\Cref{fig:reg_line} shows the regression lines learned by the model at $\tau = t$ with and without the importance weights, where the weights  ${q_1(x)}/{q_t(x)}$ are estimated by \ac{CKLIEP}. We can see that the line learned with importance weights fits $D_1$ more accurately than the one without the weights. This enables the model to make reasonable predictions on test samples from $D_1$ when the training set of $D_1$ is not  available.

\subsection{Tracing distribution shifts via KL-divergence} \label{sec:app_trace}

\begin{figure*}[t]
    \centering
    \begin{subfigure}{0.32\linewidth}
    \includegraphics[width=\linewidth,trim={0.5cm .2cm 1.5cm 1.cm},clip]{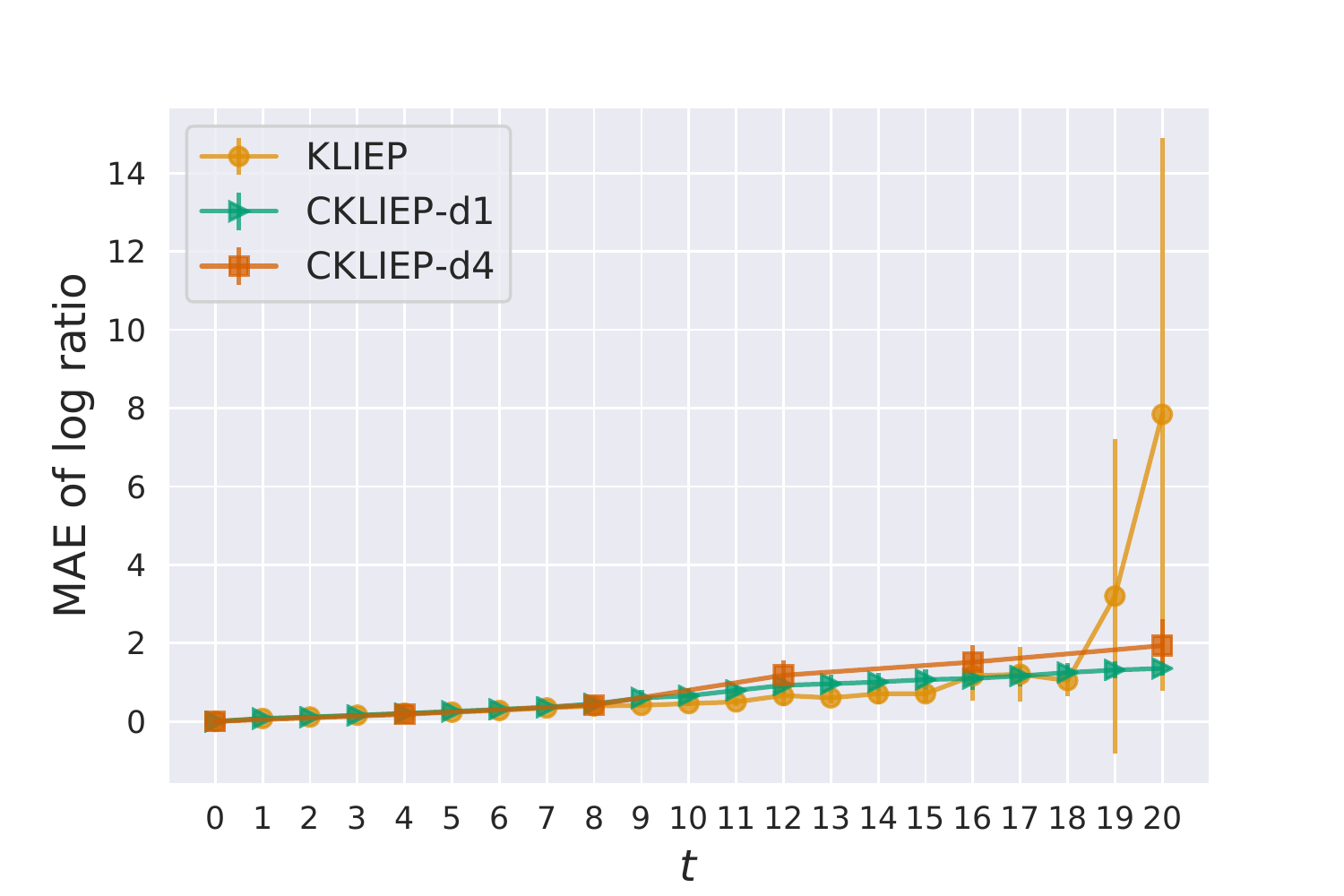}
    \caption{Comparing \acs{MAE} of log ratios estimated by \ac{CKLIEP} and \ac{KLIEP} in the scenario of a single original distribution}
    \label{fig:mae}
    \end{subfigure}
    \hfill%
    \begin{subfigure}{0.32\linewidth}
    \includegraphics[width=\linewidth,trim={0.3cm .2cm 1.5cm 1.cm},clip]{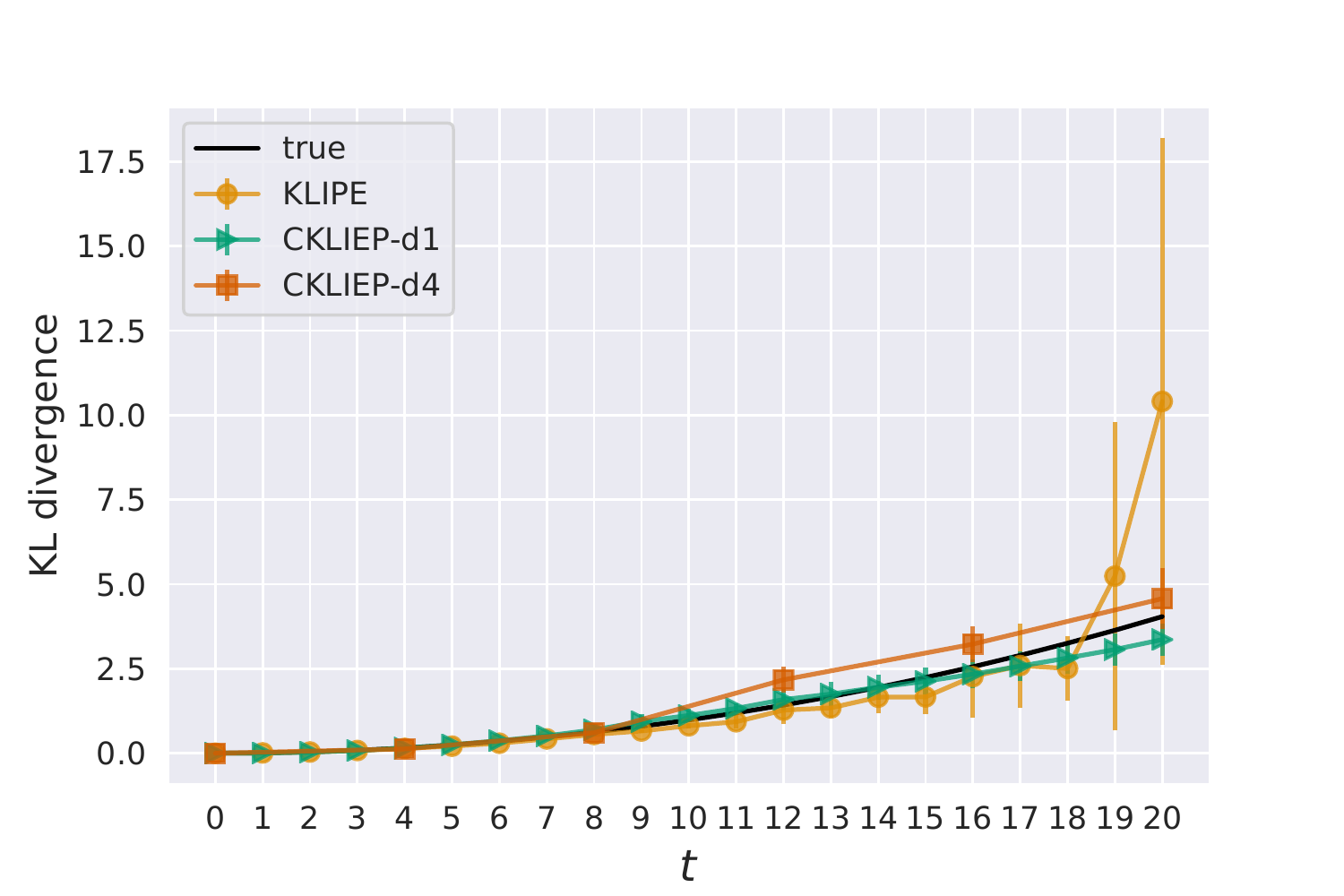}
    \caption{Comparing  KL-divergence estimated by \ac{CKLIEP} and \ac{KLIEP} in the scenario of a single original distribution}
    \label{fig:kl_single}
    \end{subfigure}
    \hfill%
    \begin{subfigure}{0.32\linewidth}
    \includegraphics[width=\linewidth,trim={0.3cm .2cm 1.5cm 1.cm},clip]{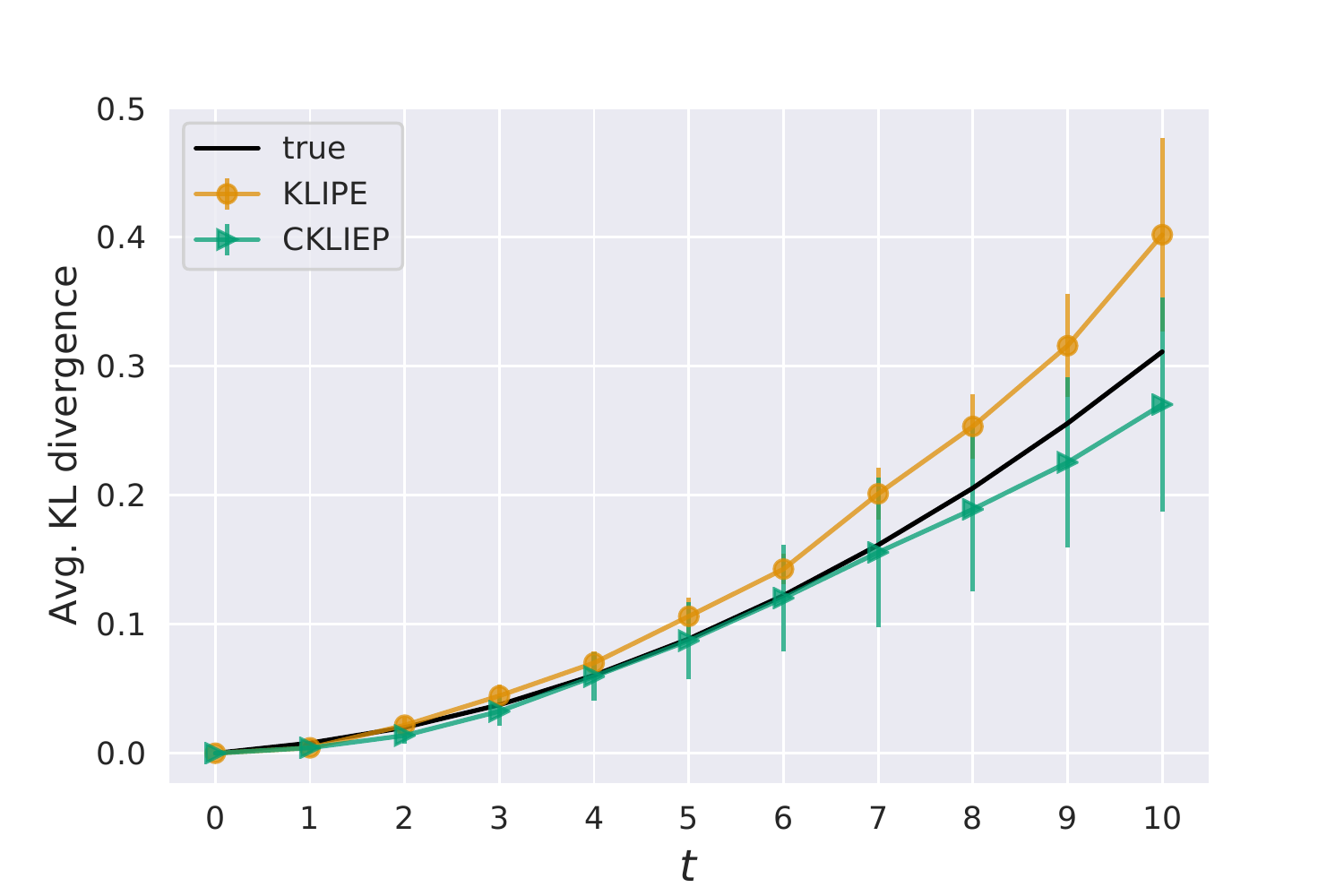}
    \caption{Comparing average KL-divergence estimated by \ac{CKLIEP} and \ac{KLIEP} in the scenario of multiple original distributions}
    \label{fig:kl_multi}
    \end{subfigure}
\caption{Comparing the performance of \ac{CKLIEP} and \ac{KLIEP} by synthetic data in the scenarios of the single and multiple original distributions. (a) \& (b) compare the \ac{MAE} of log ratios and estimated KL-divergences for a single original distribution, (c) compares the average KL-divergences for multiple original distributions. The true values of KL-divergences are computed by true ratios. The error bar is the standard deviation of 10 runs.}
\label{fig:toy_cl}
\end{figure*}

By estimating density ratios, we can approximate \fdiv{} between two distributions:
\begin{equation}
    \mathcal{D}_{f}(p||q) = \mathbb{E}_{q}\left[f\left(\frac{p(x)}{q(x)}\right)\right] \approx \frac{1}{N}\sum_{i=1}^N   f\left(r(x_i)\right)
\end{equation}
where $x_i \sim q(x)$, $f(\cdot)$  is a convex function and $f(1)=0$.  Thus, we can trace the distribution shifts by using \ac{CDRE} to  approximate the \fdiv{} between $p_{\tau}(x)$ and $q_{\tau,t}(x)$ in the online setting described in  \Cref{sec:ol_setting}. In our implementations, we apply \ac{CKLIEP} to instantiate \ac{CDRE} and choose the KL-divergence (i.e. setting $f(r)= -\log(r)$) to instantiate \fdiv{}. We compare the performance of  \ac{CKLIEP} with \ac{KLIEP} and true values using synthetic Gaussian data, where \ac{KLIEP} has access to samples of all original distributions at all time. The sample size of each distribution in the experiments is 50000. 

We first simulate the scenario of a single original distribution which is a 64-D Gaussian distribution $p(x) = \mathcal{N}(\mu_0, \sigma_0^2 I)$, where $\mu_0 = 0, \sigma_0 = 1$. At each time step, we shift the  distribution by a constant change on its mean and variance: $q_t(x) = \mathcal{N}(\mu_t, \sigma_t^2I), \mu_t=\mu_0+\Delta \mu * k, \sigma_t = \sigma_0-\Delta \sigma*k, \Delta \mu = \Delta \sigma = 0.02$, $k$ is the number of time steps within one estimation interval. We set the total number of time steps to 20. We estimate $p(x)/q_t(x)$ by applying \ac{CKLIEP} with two different time intervals: (1). \ac{CKLIEP}-d1 is to estimate $p(x)/q_t(x)$ at each time step, i.e. $k=1$; (2). \ac{CKLIEP}-d4 is to estimate $p(x)/q_t(x)$ at every four time steps, i.e. $k = 4$. We compare the \ac{MAE} of log ratios ($
    mae = \frac{1}{N}\sum_{n=1}^N |{\log r^*(x_n) - \log \hat{r}(x_n)}|
$) estimated by \ac{CKLIEP} and \ac{KLIEP} in  \Cref{fig:mae}, and compare the estimated KL-divergence with the true value in \Cref{fig:kl_single}. According to Corollary 1, the difference between $q_{t-1}(x)$ and $q_t(x)$ plays an important role in the estimation convergence, which explains why \ac{CKLIEP}-d4 gets worse performance than \ac{CKLIEP}-d1. 
\ac{KLIEP} can be viewed as a special case of \ac{CKLIEP} when  $q_{t-1}(x)=p(x)$, so the difference between $p(x)$ and $q_t(x)$ is critical to its convergence as well. We can see that in \Cref{fig:mae,fig:kl_single} \ac{KLIEP} has become much worse at the last two steps due to the two distributions are too far away from each other and thus causes serious difficulties in its convergence with a fixed sample size. 

We also simulate the scenario of multiple original distributions by 64-D Gaussian data: $p_{\tau}(x) =
\mathcal{N}(\mu_{\tau},\sigma_{\tau}^2 I)$, where $\tau \in \{1, 2, \dots, 10\}, \mu_{\tau} = 2\tau,  \sigma_{\tau} = 1$. We shift each joined original distribution ($p_{\tau}(x), \forall \tau < t$) by a constant change as the single pair scenario and set $k=1, \Delta \mu = \Delta \sigma = 0.01$, and add a new original distribution ($p_t(x)$) at each time step. In \Cref{fig:kl_multi}, we compare the averaged  KL-divergences ($\bar{\mathcal{D}} = \frac{1}{t} \sum_{\tau=1}^t \mathcal{D}_{KL}(q_{\tau,t}||p_{\tau})$) estimated by \ac{CKLIEP} and \ac{KLIEP} with the true value. \ac{CKLIEP} outperforms \ac{KLIEP} when $t$ increases, which aligns with the scenario of a single original distribution. 
\subsection{Evaluating generative models in \cl{} by \fdiv{}}
\begin{figure}[t!]
    \centering
    \includegraphics[width=0.85\linewidth,trim={0.cm 0.cm 0.cm 0.cm},clip]{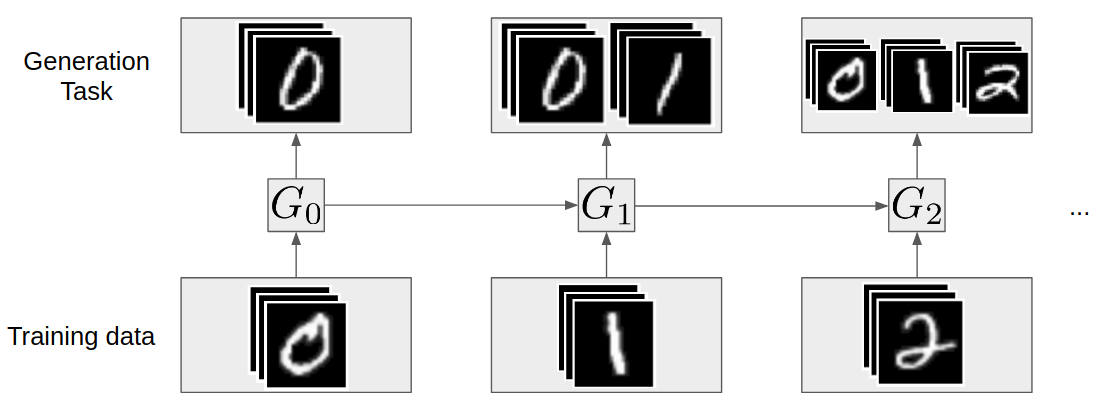}
    \caption{Demonstration of generative models in \cl{}. At task $t$
    the training set consists of samples of category $t-1$ and samples generated by the model at the previous task, and the model is to
    generate samples from all previously seen categories (figure reproduced
    from \cite{lesort2018generative}).}
    \label{fig:task_explain}
\end{figure}
\vspace{-0.1cm}
Evaluating a generative model in \cl{} can be viewed as a special case of multiple original distributions of \ac{CDRE}.
A simplified scenario for generative models in \cl{} is depicted in
\Cref{fig:task_explain}, where the goal is to learn a generative model for one
category (digit) per task while it still be able to generate samples of all previous 
categories. The training dataset of task $t$ consists of real samples of
category $t$ and samples of task $1 \cdots t-1$ generated by the previous model. 
The task index can be treated as the time index, the data distribution of the task $\tau$ can be viewed as a new original distribution $p_{\tau}(x)$ that added at time $\tau$, and its corresponding dynamic distribution $q_{\tau,t}(x)$ is the sample distribution generated by the model after trained on task $t$. The goal of generative models in \cl{} is to make $q_{\tau,t}(x)$ as close to $p_{\tau}(x)$ as possible.
In the sense of measuring the difference between $q_{\tau,t}(x)$ and $p_{\tau}(x)$, we can evaluate a generative model in \cl{} by estimating the averaged \fdiv{} over all learned tasks:
\begin{equation*}
    \bar{\mathcal{D}}_t = \frac{1}{t} \sum_{\tau=1}^t  \mathbb{E}_{q_{\tau,t}}[f(r_{\tau,t}(x))] \approx 
    \frac{1}{t} \sum_{\tau=1}^t  \sum_{n=1}^{N_\tau} f(r_{\tau,t}(x_n))
\end{equation*}
Estimating \fdiv{} using density ratios has been well studied \citep{kanamori2011f,nguyen2010estimating}. Although the density ratio can also be estimated by the density functions of distributions, explicit density functions are often not available in practice. In the absence of prior work on evaluating generative models by \fdiv{}, we provide experimental results in the supplementary material for demonstrating differences between \fdiv{}, \ac{FID}, \ac{KID}, and \ac{PRD} \citep{sajjadi2018assessing}. Through these experiments, we show that \fdiv{} can be alternative measures of generative models and one may obtain richer criteria by them.

In our experiments, 
the evaluated \acp{GAN} include \acs{WGAN} \citep{arjovsky2017wasserstein}, \acs{WGAN-GP} \citep{gulrajani2017improved}, and two members of \fgan s \citep{nowozin2016f}: \fgan-rvKL and \fgan-JS, which instantiate the \fgan{} by reverse KL and Jesen-Shannon divergences respectively. 
All \acp{GAN} are tested as conditional \acp{GAN} \citep{mirza2014conditional} using 
task indices as conditioners, and one task includes a single class of the data. We trained the \acp{GAN} on Fashion-MNIST \citep{xiao2017/online}. The sample size is 6000 for each class. More details of the experimental settings and the extra results with MNIST \citep{lecun2010mnist} are provided in the supplementary material. We evaluate these  \acp{GAN} in \cl{} by a few members of \fdiv{} which are estimated by \ac{CKLIEP},  
and compare the results with common measures \ac{FID}, \ac{KID} as used in static learning. 

\begin{figure}[t!]
    \centering
    \begin{subfigure}{0.485\linewidth}
    \includegraphics[width=\linewidth,trim={.2cm .1cm .7cm 1.4cm},clip]{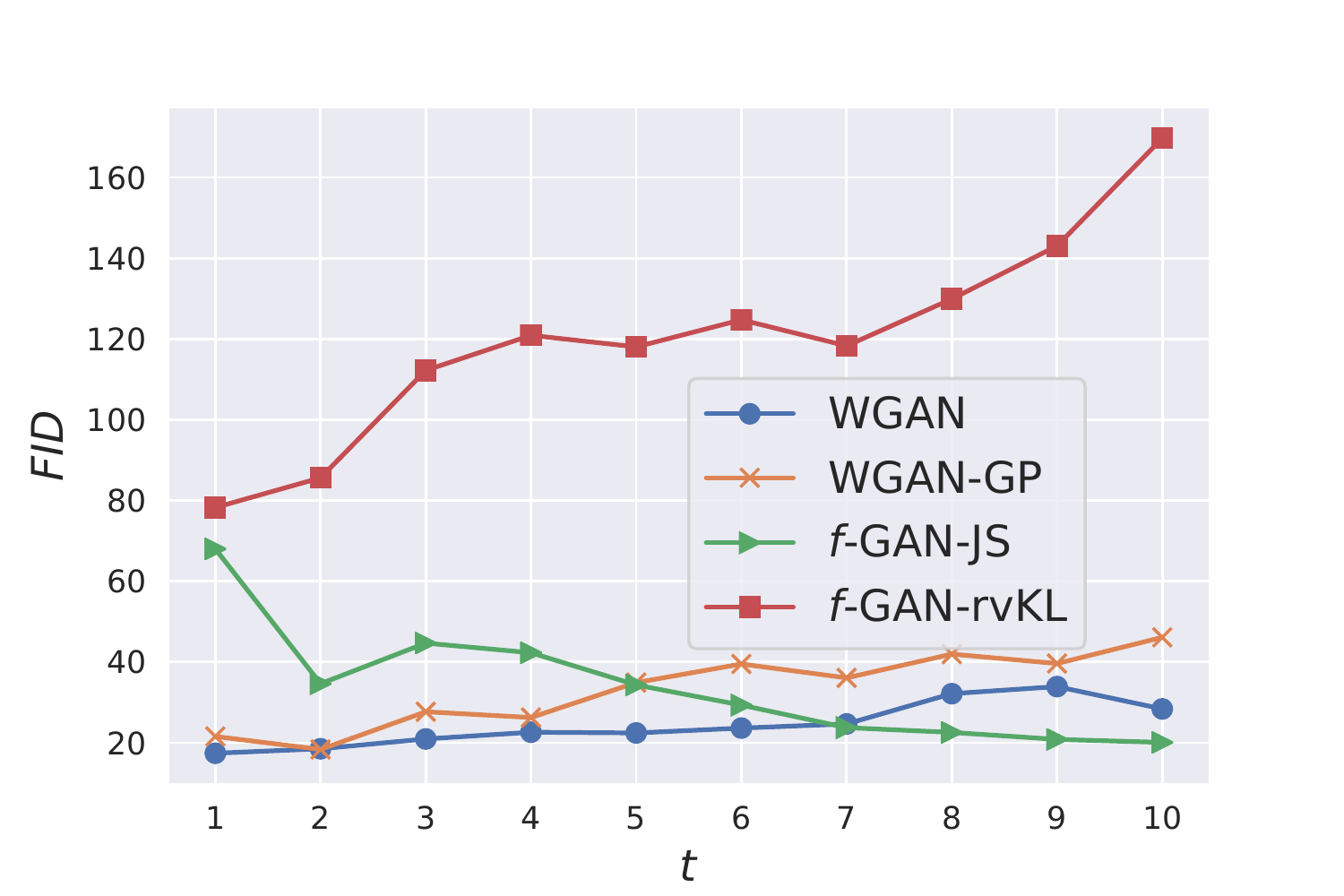}
    \caption{FID}
    \label{fig:fashion_fid}
    \end{subfigure} 
   \hfill %
    \begin{subfigure}{0.485\linewidth}
    \includegraphics[width=\linewidth,trim={.2cm .1cm .7cm 1.4cm},clip]{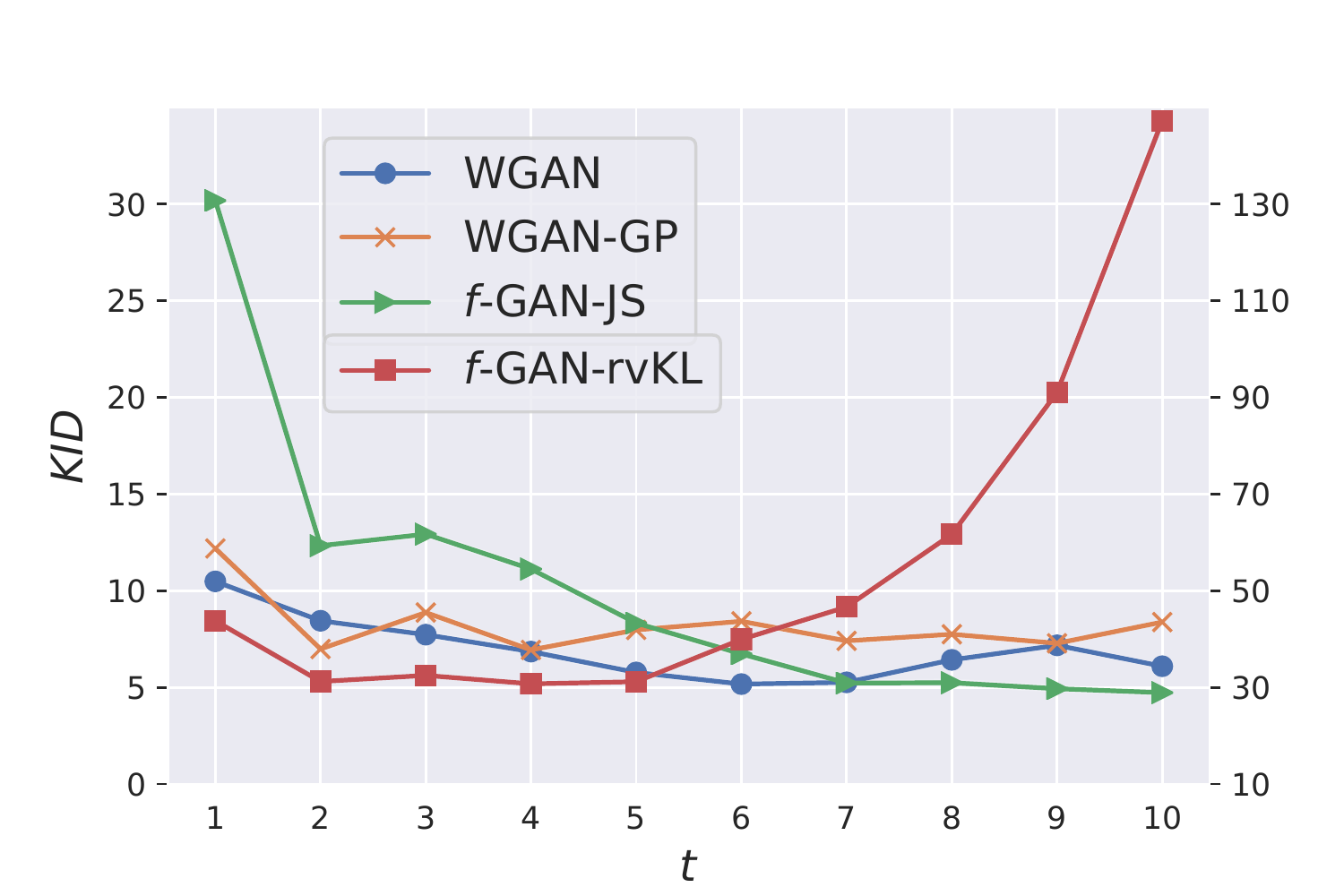}
    \caption{KID}
    \label{fig:fashion_kid}
    \end{subfigure}
    \\%
    \begin{subfigure}{0.485\linewidth}
    \includegraphics[width=\linewidth,trim={.2cm .1cm 1.cm 1.4cm},clip]{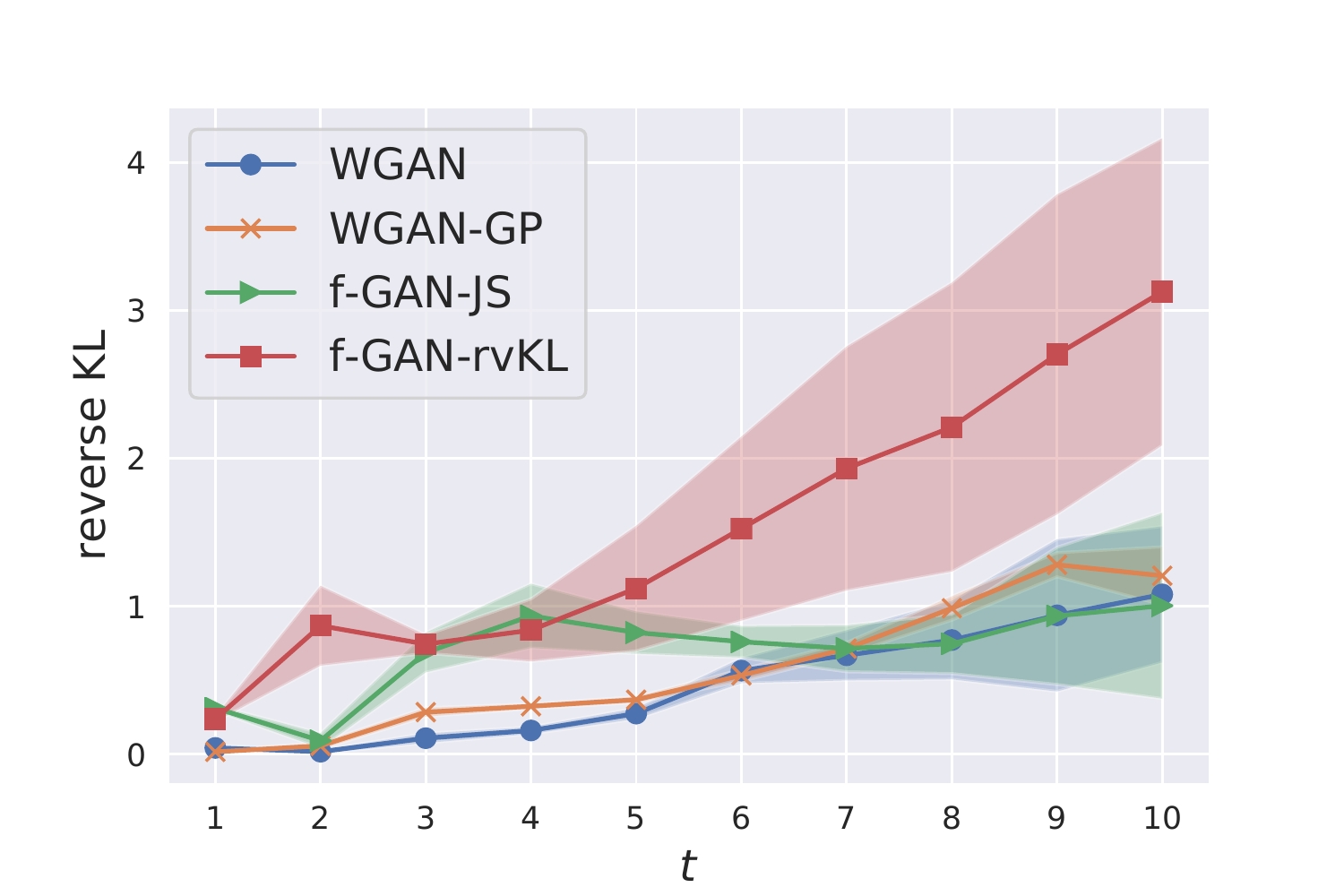}
    \caption{CKLIEP (reverse KL)}
    \label{fig:fashion_klqp}
    \end{subfigure}
    \hfill%
    \begin{subfigure}{0.485\linewidth}
    \includegraphics[width=\linewidth,trim={0.2cm 0.1cm 1.cm 1.4cm},clip]{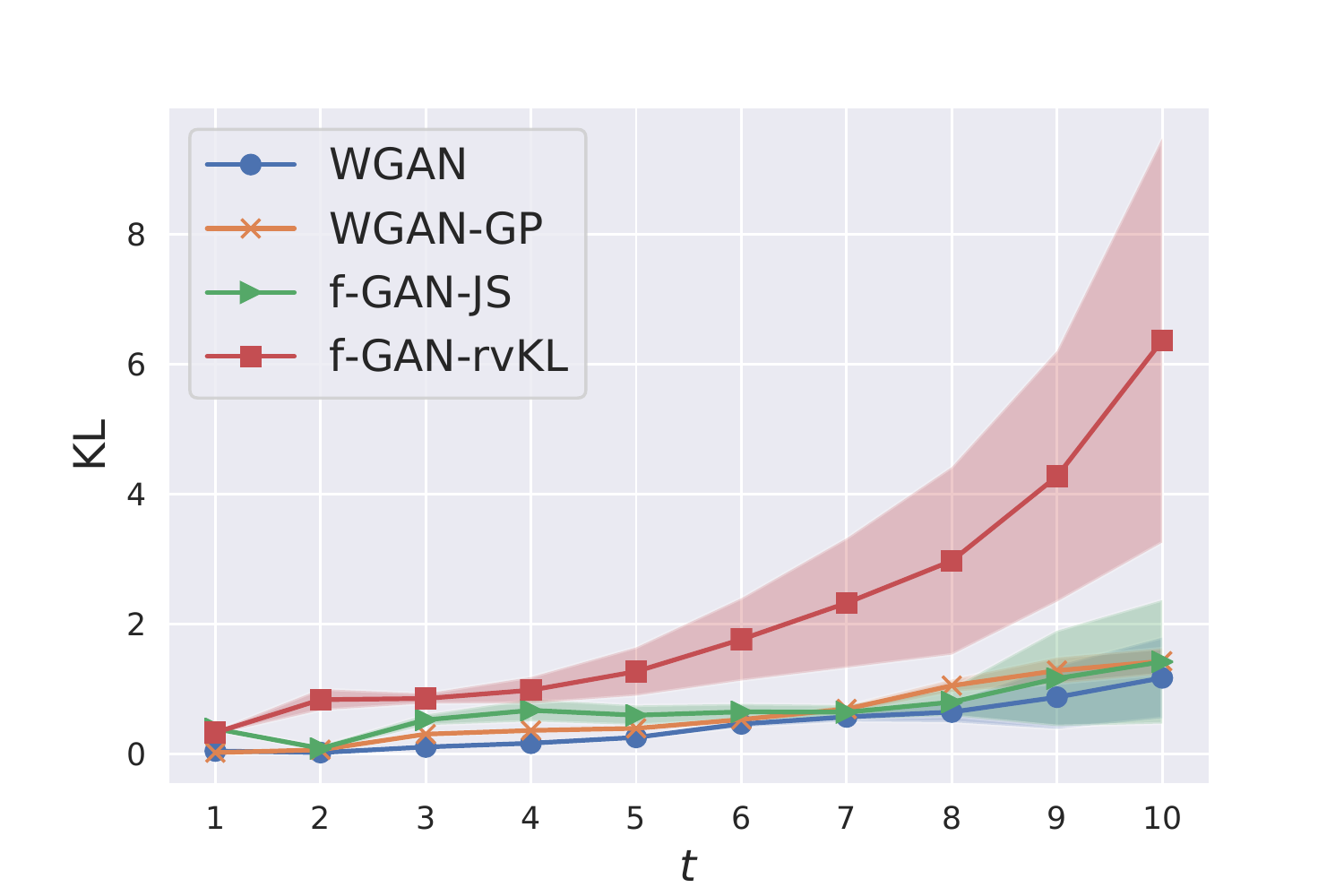}
    \caption{CKLIEP (KL)}
    \label{fig:fashion_klpq}
    \end{subfigure}
    \\%
    \begin{subfigure}{0.485\linewidth}
    \includegraphics[width=\linewidth,trim={.2cm .1cm 1.cm 1.4cm},clip]{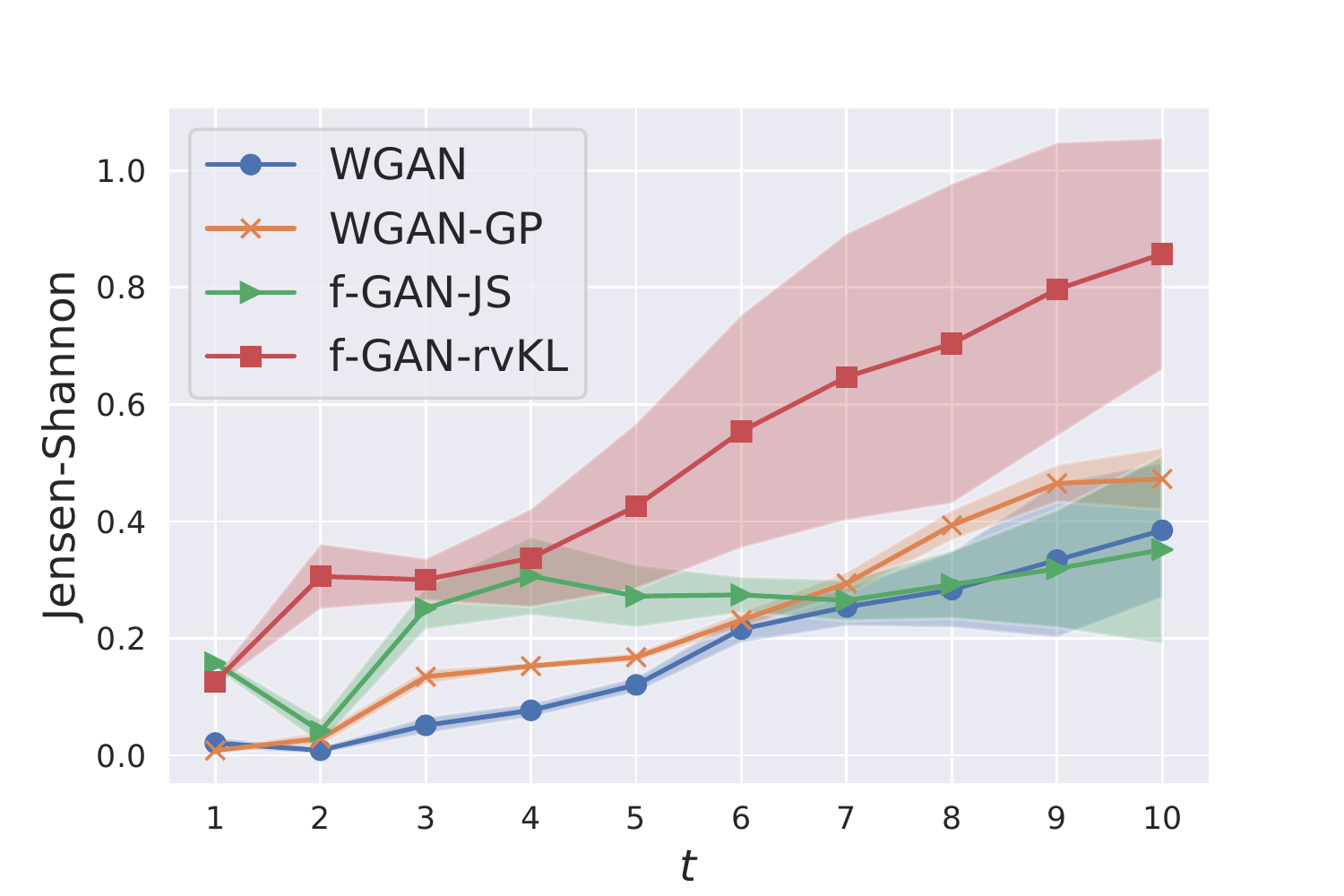}
    \caption{CKLIEP (JS)}
    \label{fig:fashion_js}
    \end{subfigure}
    \hfill%
    \begin{subfigure}{0.485\linewidth}
        \includegraphics[width=\linewidth,trim={.2cm .1cm 1.cm 1.4cm},clip]{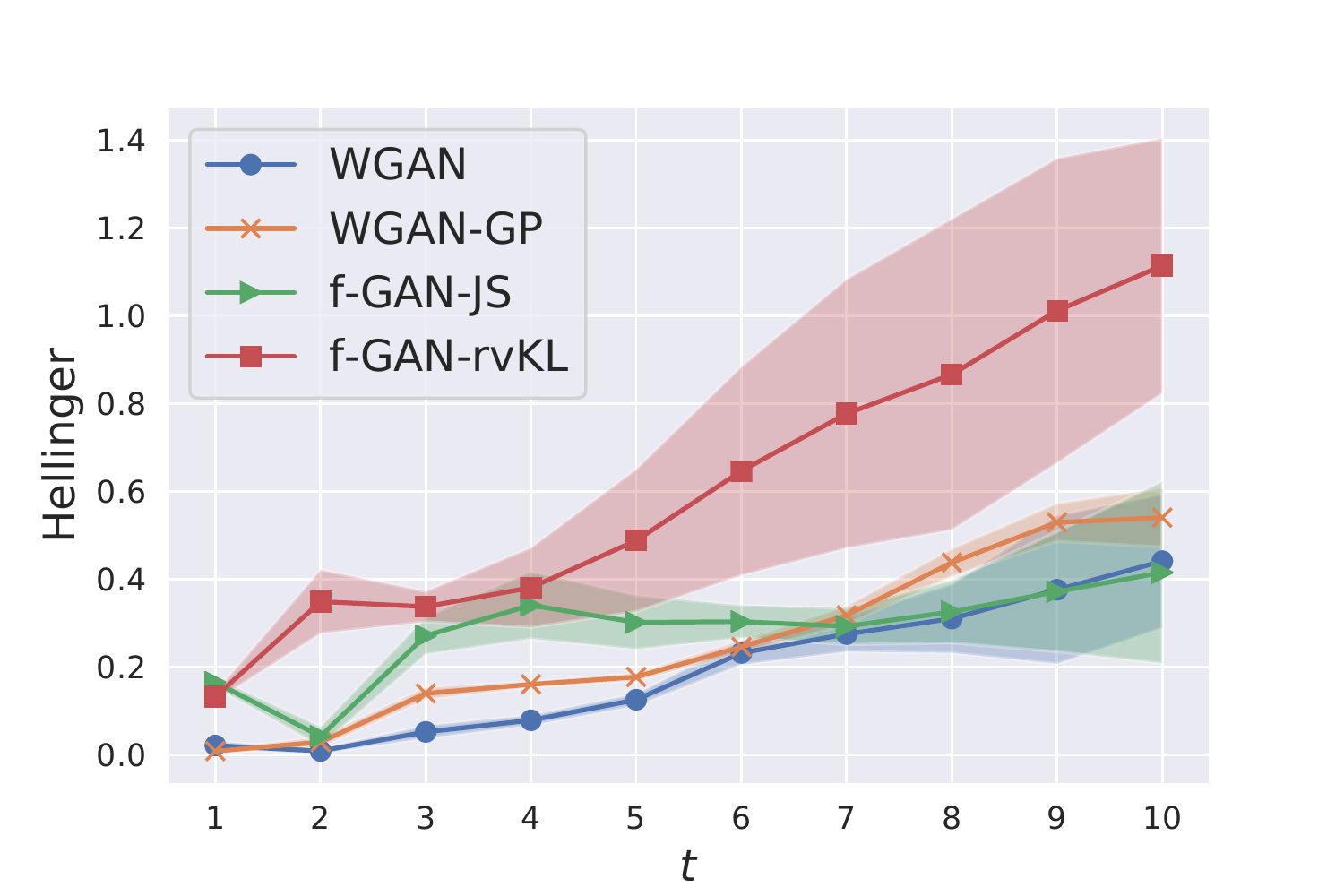}
        \caption{CKLIEP (Hellinger)}
        \label{fig:fashion_hg}
        \end{subfigure}
    \caption{Evaluating GANs in \cl{} on Fashion-MNIST. The shaded area
    are plotted by standard deviation of 10 runs. The x-axis is task index and y-axis is the specified measurement as in each sub-caption. The y-axis in the right side of \Cref{fig:fashion_kid} is the y-axis of the red line (\fgan-rvKL), which is in a much larger scale than others. All the measures are the lower the better.}
    \label{fig:fashion_cl}
\end{figure}

\begin{figure}[t!]
    \centering
    \begin{subfigure}{0.45\linewidth}
    \includegraphics[width=\linewidth,trim={3.cm 3.cm 2.cm 2.5cm},clip]{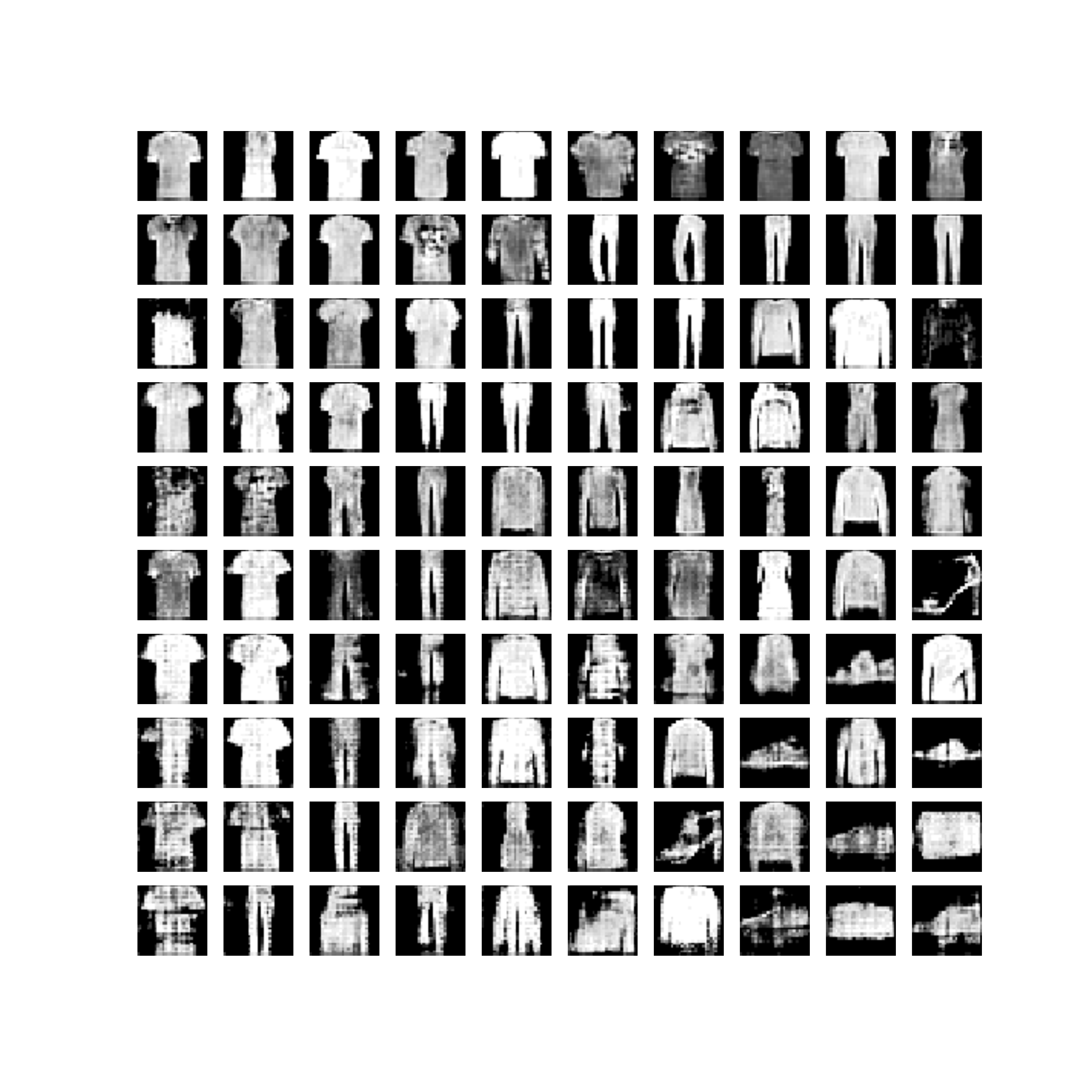}
    \caption{WGAN}
    \label{fig:wgan_fashion}
    \end{subfigure}
    \quad%
    \begin{subfigure}{0.45\linewidth}
    \includegraphics[width=\linewidth,trim={3.cm 3.cm 2.cm 2.5cm},clip]{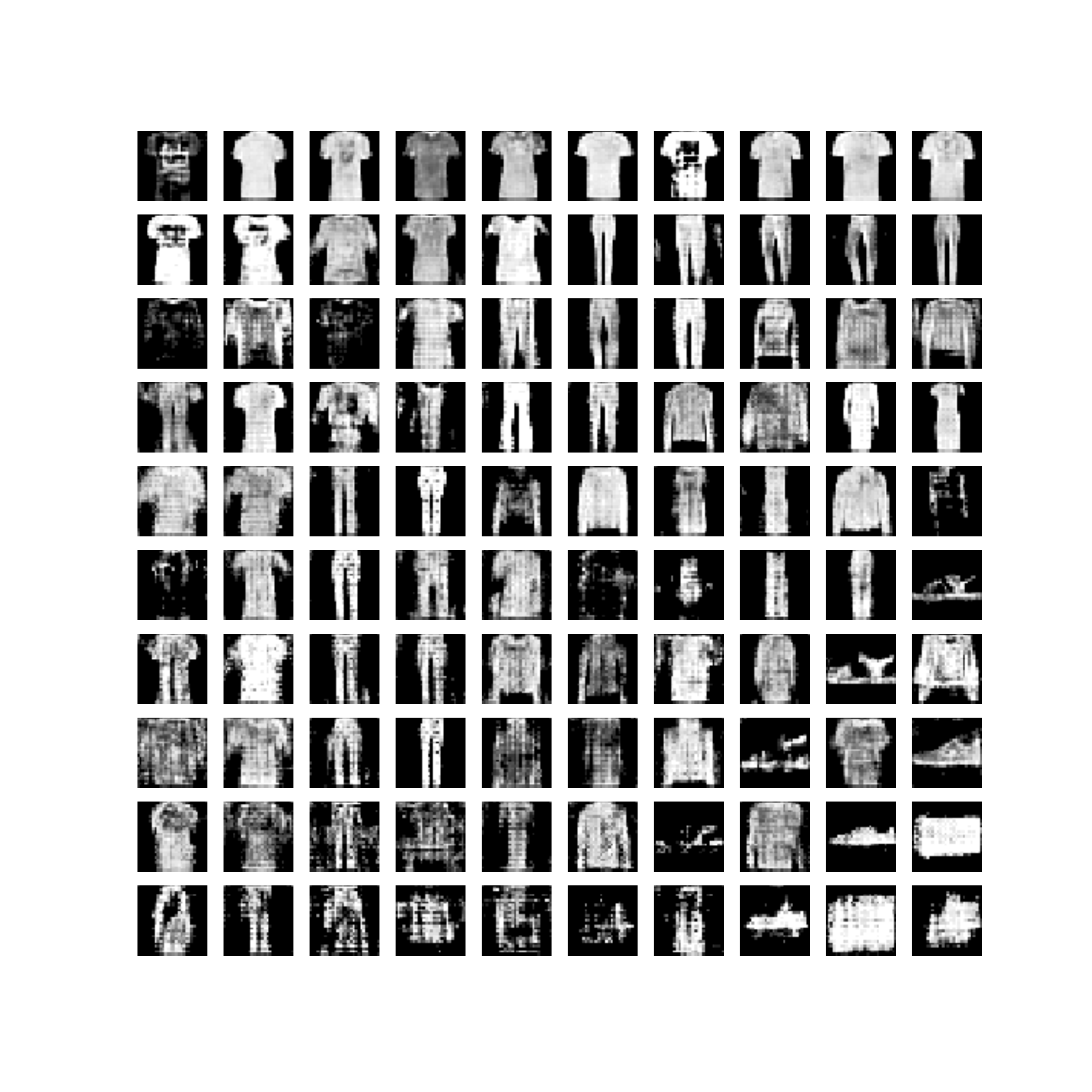}
    \caption{WGAN-GP}
    \label{fig:wgangp_fashion}
    \end{subfigure}
    \\%
    \begin{subfigure}{0.45\linewidth}
    \includegraphics[width=\linewidth,trim={3.cm 3.cm 2.cm 2.5cm},clip]{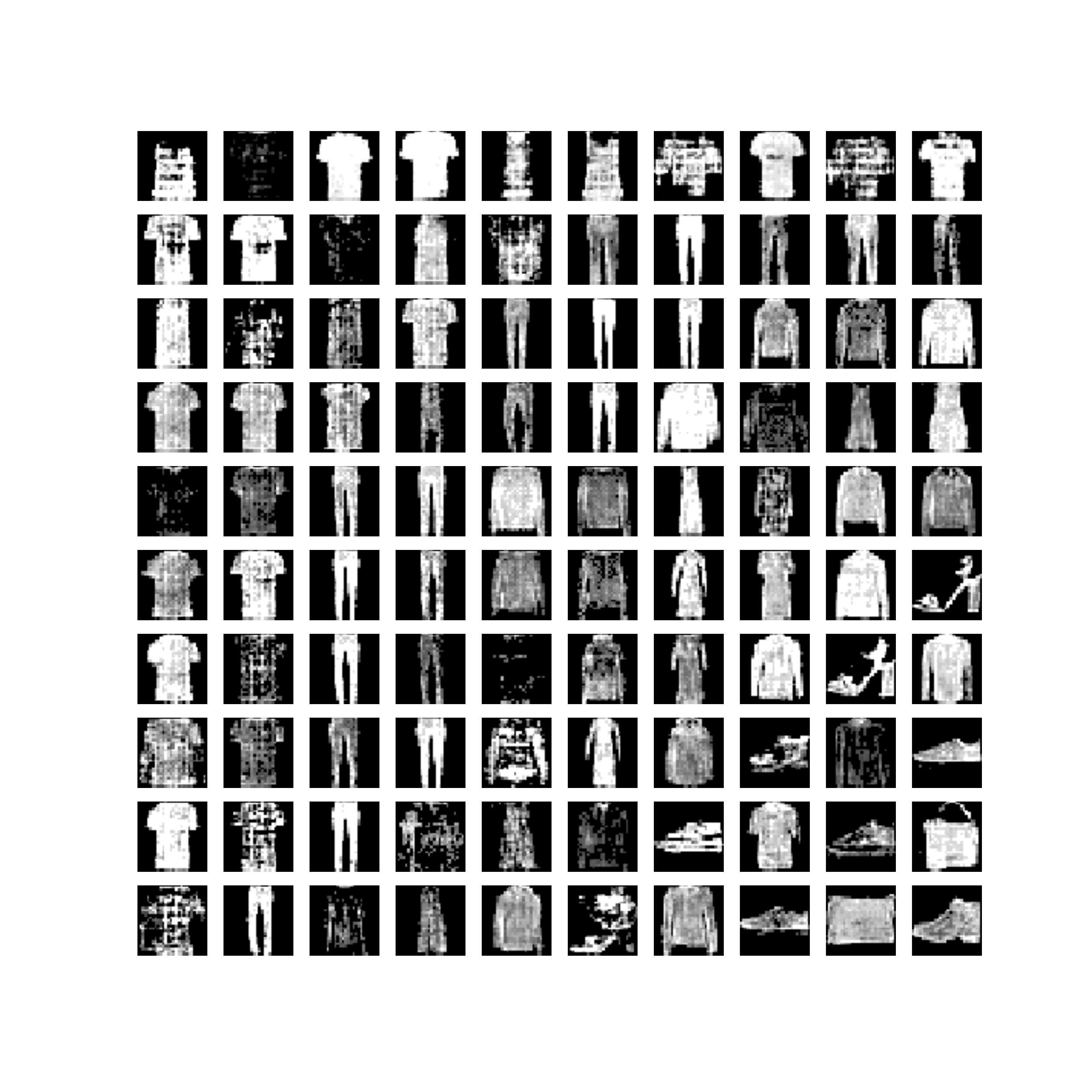}
    \caption{\fgan-JS}
    \label{fig:js_fashion}
    \end{subfigure}
    \quad%
    \begin{subfigure}{0.45\linewidth}
    \includegraphics[width=\linewidth,trim={3.cm 3.cm 2.cm 2.5cm},clip]{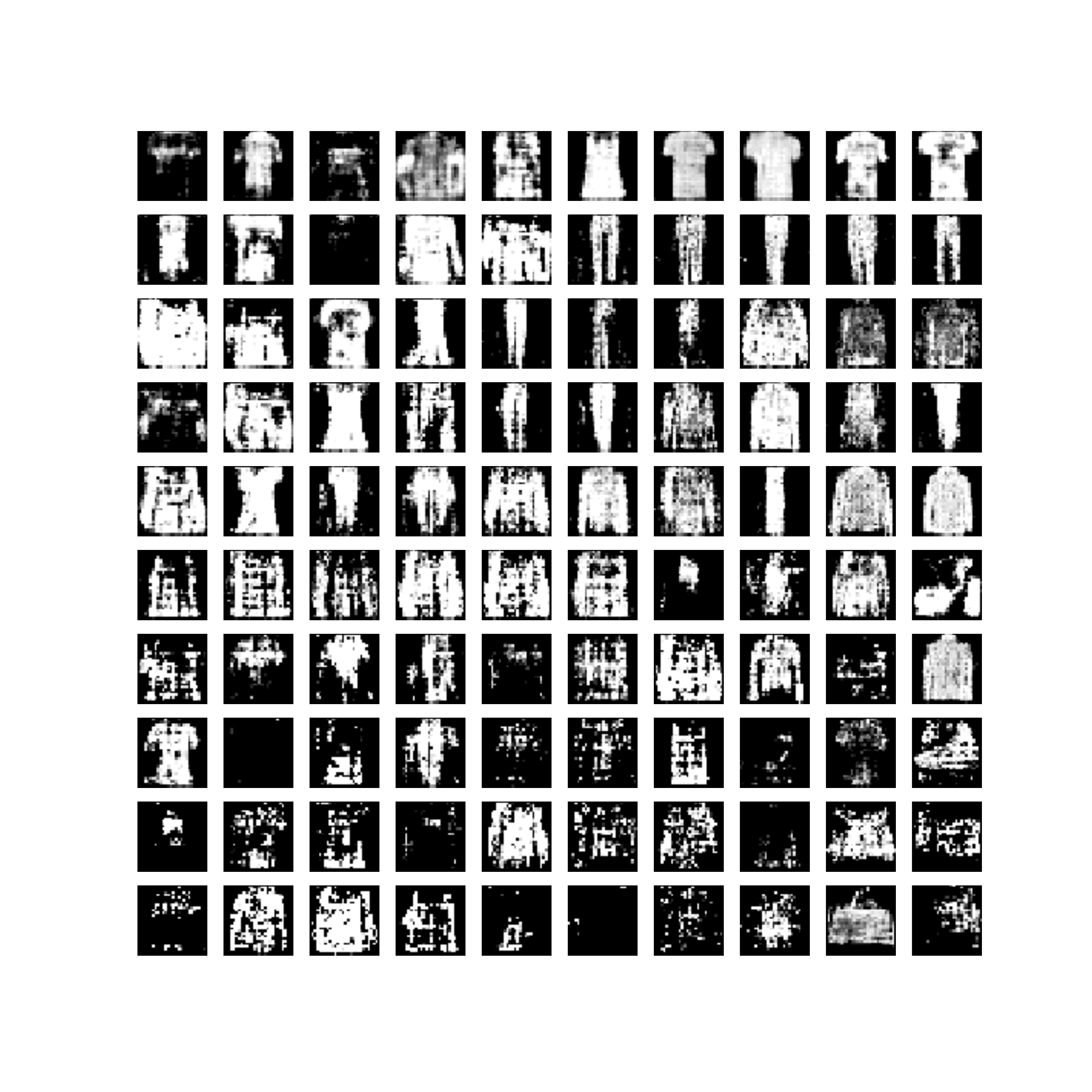}
    \caption{\fgan-rvKL}
    \label{fig:rvkl_fashion}
    \end{subfigure}
    \caption{Fashion-MNIST samples generated by several  \acp{GAN} in \cl{}. In each sub-figure, each row displays images generated by the model at each task, the order is from the top to bottom (task 1 to 10). The generated  samples are from all learned classes at task $i$. The displayed samples are uniformly  randomly chosen from generated samples of each class.}
    \label{fig:fashion_task_samples}
\end{figure}

We deployed two feature generators in the  experiments: 1) A classifier pre-trained on real samples of all classes, which is a \ac{CNN} and the extracted features are the activations of the last hidden layer (similar with inception feature); 2) A \ac{CVAE} trained along with the procedure of \cl{}, 
and the features are the output of the encoder. The dimension of features are 64 for both  classifier and \ac{CVAE}. We deployed the pre-trained classifier as the feature generator for \ac{FID},\ac{KID}, and deployed the \ac{CVAE} as the feature generator for \ac{CKLIEP} in all experiments. 
\Cref{fig:fashion_cl} compares the evaluations for the \acp{GAN} by \ac{FID}, \ac{KID}, and four members of \fdiv{} (which are estimated by \ac{CKLIEP}): KL, reverse KL, Jensen-Shannon, Hellinger.  All these measures are the lower the better. We also display randomly chosen samples generated by those \acp{GAN} in \Cref{fig:fashion_task_samples} for a better understanding of the evaluations.   

In general, all measurements give similar evaluations. For example, \fgan-rvKL has the worst performance on all measures during the whole task sequence, whereas \acs{WGAN} and \acs{WGAN-GP} have similar performance on all measures. 
One main disagreement is that \fgan-JS shows a decreasing trend on \ac{FID} and \ac{KID} but shows a slightly increasing trend on members of \fdiv{}. According to displayed samples of \fgan-JS (\Cref{fig:js_fashion}),  there is no notable improvement observed while \fgan-JS learning more tasks, hence, the evaluations given by \ac{FID} and \ac{KID} are  doubtful in this case.
Moreover, \ac{KID} also shows doubtful evaluations for \acs{WGAN} and \acs{WGAN-GP}. 
Visually, samples from \acs{WGAN} and \acs{WGAN-GP} (\Cref{fig:wgan_fashion,fig:wgangp_fashion}) are obviously losing fidelity while learning more tasks, which matches the increasing trend in all measures except \ac{KID}. In principle, \fdiv{} may have different opinions with \ac{FID} and \ac{KID} because \fdiv{} are based on density ratios which may give more attention on parts with less probability mass (due to the ratio of two small values can be very large) whereas  
\ac{FID} and \ac{KID} are based on \ac{IPM} \citep{sriperumbudur2012empirical} which focus on parts of the distribution with most probability mass. We illustrate this by a demo experiment in the supplementary material. 
All in all, these experimental results demonstrate that \fdiv{} estimated by \ac{CKLIEP} can provide meaningful evaluations for generative models in \cl{}.

\section{Conclusion and discussion}\label{sec:discuss}
We proposed a novel method \ac{CDRE} for estimating density ratios in an online setting. We demonstrate the efficacy of our method in a range of online applications, such as tracing distribution shifts by KL-divergence, backwards covariate shift, evaluating generative models in continual learning.
The experiments showed that it can obtain better performance than common \ac{DRE} methods without the need of storing historical samples in the streaming data.   
We also demonstrated that \ac{CDRE} can provide  an alternative approach for model selection in \cl{} when other measures are not applicable. In addition, we proposed a simple approach \ac{CVAE} for feature generation in \cl{} when a pre-trained classifier is not available. Our experiments showed that \ac{CDRE} combined with \ac{CVAE} can work well on high-dimensional data, which is a difficult scenario for density ratio estimation. 

When estimating divergences which are not based on log-ratios it may be better try some other form of ratio estimators (\emph{i.e.} replacing \ac{KLIEP}'s formulation). For instance, it may be preferable to use \ac{LSIF} \citep{kanamori2009efficient} when estimating Pearson $\chi^2$ divergence, since a small deviation in log-ratio can result in large squared errors. Also, since \ac{LSIF} itself is based on Pearson $\chi^2$ divergence, it appears to be a more natural choice. On the other hand, it has been discussed in \citet{mohamed2016learning} that the discriminator of some type of \acp{GAN} can be viewed as a density ratio estimator. For instance, it is feasible to apply the formulation of the discriminators of \fgan{} \citep{nowozin2016f} to the basic estimator in \ac{CDRE}. We plan to explore further possibilities of \ac{CDRE}. 
It is also possible to estimate the Bregman divergence by ratio estimation \citep{uehara2016generative}, giving even more options for the online applications of \ac{CDRE}.
\ac{DRE} also has many other
applications other than estimating divergences, such as change detection
\citep{liu2013change} and mutual information estimation \citep{sugiyama2012density}. Likewise, \ac{CDRE} may be useful for more applications in the online setting which we
would like to explore in the future.  Finally, we mention \ac{CDRE}'s potential to estimating ratios when the difference between the two distributions is significant by accompanying with methods for sampling intermediate distributions, e.g. the one introduced in \cite{rhodes2020telescoping}.

%% file: supp.tex
\appendix

\section{Proof of asymptotic normality of \acs{CKLIEP}}\label{sec:proof}

Define $\hat{\beta}_t$ as the estimated parameter that satisfies: 
\begin{equation}\label{eq:def_hat}
    \mathcal{L}^'_t(\hat{\beta}_t) \triangleq \nabla_{\beta_t} \mathcal{L}_t(\beta_t)\big|_{\beta_t = \hat{\beta}_t} =   0
\end{equation}

 Assume $\phi_{\beta_t}(x)$ ($\phi_{\beta_t}(x) = \psi_{\beta_t}(x) - \psi_{\beta_{t-1}}(x)$) includes the correct function that there exists $\beta_{t}^*$ recovers the true ratio over the  population:
\begin{equation}\label{eq:def_optm}
    \begin{split}
     & r^*_{s_t}(x)  
    = \frac{q_{t-1}(x)}{q_t(x)}
    = \frac{\exp\{\phi_{\beta^*_t}(x)\}}{\mathbb{E}_{ q_t}[\exp\{\phi_{\beta^*_t}(x)\}]}, \\
    & \text{where} \ \ \phi_{\beta^*_t}(x) = \psi_{\beta_t^*}(x) - \psi_{{\beta}_{t-1}}(x),
    \end{split}
\end{equation}

\textbf{\emph{Notations}}: $\leadsto$ and $\xrightarrow{P}$ mean convergence in distribution and convergence in probability, respectively. 

\textbf{\emph{Assumptions}}: We assume $q_t(x)$ and $q_{t-1}(x)$ are independent, $n_t = n_{t-1} = n$, where $n_t$ is the sample size of $q_t(x)$. Let $S_t$ be the support of $q_t$, we assume $S_{t-1} \subseteq S_t$ in all cases. 

\begin{lemma}\label{lemma1}
  Let $\ell^'_r(\beta^*_t) \triangleq \frac{1}{n}\sum_{j=1}^{n}\nabla_{\beta_t} \log r_{s_t}(x_{t-1,j})|_{\beta_t = \beta^*_t}$, we have   $\sqrt{n}\ell^'_r(\beta^*_t) \leadsto \mathcal{N}(0, \sigma^2)$, where   
\begin{equation*}
\begin{split}
 \sigma^2 &= Cov_{q_{t-1}}[\nabla_{\beta_t}\phi_{\beta^*_t}(x)] + 
 \frac{Cov_{q_t}[\nabla_{\beta_t}\exp\{\phi_{\beta^*_t}(x)\}]}{\mathbb{E}_{q_t}[\exp\{\phi_{\beta^*_t}(x)\}]^2}
\end{split}
\end{equation*}
\end{lemma}

\begin{proof}
Because 
\begin{equation}\label{eq:dlogr}
\begin{split}
    &\nabla_{\beta_t} \log r_{s_t}(x) = \nabla_{\beta_t}\phi_{\beta_t}(x) - \frac{\sum_{i}^{n}\nabla_{\beta_t}\exp\{\phi_{\beta_t}(x_{t,i})\}}{ \sum_{i}^{n} \exp\{\phi_{\beta_t}(x_{t,i})\} }
\end{split}
\end{equation}
then 
\begin{equation}\label{eq:lr}
    \begin{split}
      \sqrt{n}\ell^'_r(\beta_t) 
      &= \frac{\sqrt{n}}{n}\sum_{j=1}^{n}\nabla_{\beta_t}\phi_{\beta_t}(x_{t-1,j}) - \\
      & \quad
      \sqrt{n} \frac{\frac{1}{n}\sum_{i}^{n}\nabla_{\beta_t}\exp\{\phi_{\beta_t}(x_{t,i})\}}{ \frac{1}{n}\sum_{i}^{n} \exp\{\phi_{\beta_t}(x_{t,i})\} }  
    \end{split}
\end{equation}
By the central limit theorem we have: 
\begin{equation}
    \begin{split}
    & \frac{1}{n}\sum_{j=1}^{n}\nabla_{\beta_t}\phi_{\beta_t}(x_{t-1,j}) \leadsto \mathcal{N}\left(\mathbb{E}_{q_{t-1}}[\nabla_{\beta_t}\phi_{\beta_t}(x)], \frac{Cov_{q_{t-1}}[\nabla_{\beta_t}\phi_{\beta_t}(x)]}{n}\right), \\ 
    & \frac{1}{n}\sum_{i}^{n}\nabla_{\beta_t}\exp\{\phi_{\beta_t}(x_{t,i})\} \leadsto \\
    & \qquad \qquad 
    \mathcal{N} \left(\mathbb{E}_{q_t}[\nabla_{\beta_t}\exp\{\phi_{\beta_t}(x)\}], \frac{Cov_{q_t}[\nabla_{\beta_t}\exp\{\phi_{\beta_t}(x)\}]}{n}] \right), \\
    \end{split}
\end{equation}
and by the weak law of large numbers:
\begin{equation}
    \begin{split}
     &
    \frac{1}{n}\sum_{i}^{n} \exp\{\phi_{\beta_t}(x_{t,i})\} \xrightarrow{P} \mathbb{E}_{q_t}[\exp\{\phi_{\beta_t}(x)\}]  
    \end{split}
\end{equation}

Because $q_t(x)$ and $q_{t-1}(x)$ are assumed independent,
 combine the above results we get: 
\begin{equation}
    \begin{split}
        &\sqrt{n}\ell^'_r(\beta^*_t) \leadsto \mathcal{N}(\mu, \sigma^2), \\
        &\mu = \sqrt{n}\left(\mathbb{E}_{q_{t-1}}[\nabla_{\beta_t}\phi_{\beta^*_t}(x)] - \frac{\mathbb{E}_{q_t}[\nabla_{\beta_t}\exp\{\phi_{\beta^*_t}(x)\}]}{\mathbb{E}_{q_t}[\exp\{\phi_{\beta^*_t}(x)\}]}\right), \\
        & \sigma^2 = Cov_{q_{t-1}}[\nabla_{\beta_t}\phi_{\beta^*_t}(x)] +  \frac{Cov_{q_t}(\nabla_{\beta_t}\exp\{\phi_{\beta^*_t}(x)\})}{\mathbb{E}_{q_t}[\exp\{\phi_{\beta^*_t}(x)\}]^2}
    \end{split}
\end{equation}
Taking derivatives from both sides of $1 = \int r^*_{s_t}(x)q_t(x)dx $:
\begin{equation}\label{eq:mean_0}
\begin{split}
    0 &=  \nabla_{\beta_t}\mathbb{E}_{q_{t}}[ r^*_{s_t}(x)]
    =
     \int \nabla_{\beta_t} r^*_{s_t}(x) q_t(x) dx \\
    &= \int \frac{\nabla_{\beta_t}r^*_{s_t}(x)}{r^*_{s_t}(x)} r^*_{s_t}(x) q_t(x) dx \\
    &=  
    \int \nabla_{\beta_t}\log r^*_{s_t}(x) q_{t-1}(x) dx \\
    &= 
    \mathbb{E}_{ q_{t-1}}[\nabla_{\beta_t} \log r^*_{s_t}(x)] \\
    &=
    \mathbb{E}_{ q_{t-1}}[\nabla_{\beta_t}\phi_{\beta^*_t}(x)] -  \frac{\mathbb{E}_{q_t}[\nabla_{\beta_t}\exp\{\phi_{\beta^*_t}(x)\}]}{ \mathbb{E}_{q_t}[ \exp\{\phi_{\beta^*_t}(x)\}]}
\end{split}
\end{equation}
which gives $\mu = 0$. We prove the lemma.
\end{proof}

\begin{lemma}\label{lemma2}
Let $\ell^{''}_r(\beta^*_t) \triangleq \frac{1}{n} \sum_{j=1}^{n} \nabla^2_{\beta_t}\log r_{s_t}(x_{t-1,j})|_{\beta_t=\beta^*_t}$, we have  $\ell^{''}_r(\beta^*_t) \xrightarrow{P} -I_{\beta^*_t}$, where ${I}_{\beta^*_t} \triangleq Cov_{q_{t-1}}
[\nabla_{\beta_t}\phi_{\beta^*_t}(x)]$.
\end{lemma}

\begin{proof}
According to \Cref{eq:dlogr}
\begin{equation}
    \begin{split}
    &(\nabla_{\beta_t} \log r_{s_t}(x))^2 
    = \left(\frac{\sum_{i}^{n}\nabla_{\beta_t}\exp\{\phi_{\beta_t}(x_{t,i})\}}{ \sum_{i}^{n} \exp\{\phi_{\beta_t}(x_{t,i})\}}\right)^2
     - \\
    & \quad 2\nabla_{\beta_t} \phi_{\beta_t}(x) \frac{\sum_{i}^{n}\nabla_{\beta_t}\exp\{\phi_{\beta_t}(x_{t,i})\}}{ \sum_{i}^{n} \exp\{\phi_{\beta_t}(x_{t,i})\}} 
     + (\nabla_{\beta_t} \phi_{\beta_t}(x))^2
    \end{split}
\end{equation}
By the law of large numbers,
\begin{equation}
    \begin{split}
    &\frac{1}{n}\sum_{j=1}^{n}(\nabla_{\beta_t} \log r_{s_t}(x_{t-1,j}))^2 
    \xrightarrow{P} 
    \left(\frac{\mathbb{E}_{q_t}[\nabla_{\beta_t}\exp\{\phi_{\beta_t}(x_{t,i})\}]}{\mathbb{E}_{q_t}[\exp\{\phi_{\beta_t}(x_{t,i})\}]}\right)^2
    \\
    & \quad 
    - 2 \mathbb{E}_{q_{t-1}}[\nabla_{\beta_t} \phi_{\beta_t}(x)] \frac{\mathbb{E}_{q_t}[\nabla_{\beta_t}\exp\{\phi_{\beta_t}(x_{t,i})\}]}{\mathbb{E}_{q_t}[\exp\{\phi_{\beta_t}(x_{t,i})\}]} \\
    & \quad 
    + \mathbb{E}_{q_{t-1}}[(\nabla_{\beta_t} \phi_{\beta_t}(x))^2]
    \end{split}
\end{equation}
Substituting \Cref{eq:mean_0} to the right side of the above equation, we can get:
\begin{equation}\label{eq:lm2b}
    \begin{split}
    &\frac{1}{n}\sum_{j=1}^{n}(\nabla_{\beta_t} \log r_{s_t}(x_{t-1,j}))^2|_{\beta_t = \beta^*_t} 
    \xrightarrow{P} \\
    & \qquad  \qquad 
    \mathbb{E}_{ q_{t-1}}[(\nabla_{\beta_t} \phi_{\beta^*_t}(x))^2] - \mathbb{E}_{ q_{t-1}}[\nabla_{\beta_t}\phi_{\beta^*_t}(x)]^2 \\
    & \qquad  \qquad   
    = 
    Cov_{q_{t-1}}[\nabla_{\beta_t}\phi_{\beta^*_t}(x)] = {I}_{\beta^*_t} 
    \end{split}
\end{equation}

Because  
\begin{equation}
    \begin{split}
    &\nabla^2_{\beta_t}\log r_{s_t}(x) = \frac{\nabla^2_{\beta_t}r_{s_t}(x)}{r_{s_t}(x)} - (\nabla_{\beta_t}\log r_{s_t}(x))^2 , 
    \end{split}
\end{equation}
then according to \Cref{eq:lm2b}
\begin{equation}
    \begin{split}
    & \ell^{''}_r(\beta^*_t) \xrightarrow{P} \mathbb{E}_{q_{t-1}}\left[\frac{\nabla^2_{\beta_t}r^*_{s_t}(x)}{r^*_{s_t}(x)}\right] - I_{\beta^*_t} \\
    & \qquad \qquad  = 
      \int {\nabla^2_{\beta_t}r^*_{s_t}(x)} q_{t}(x) dx - I_{\beta^*_t}
    \end{split}
\end{equation}

Under mild assumptions we can interchange the integral and derivative operators:
\begin{equation}
  \begin{split}
      \int {\nabla^2_{\beta_t}r^*_{s_t}(x)} q_{t}(x) dx 
      &= \nabla^2_{\beta_t} \int {r^*_{s_t}(x)} q_{t}(x) dx \\
      & = \nabla^2_{\beta_t} \int \frac{q_{t-1}(x)}{q_t(x)} q_{t}(x) dx
      = 0
  \end{split}  
\end{equation}
then we prove this lemma. 
\end{proof}

\begin{lemma}\label{lemma3}
Let $\ell_c(\beta_t) \triangleq \lambda_c \left(\frac{\Psi_t(x_t)}{\Phi_t(x_t)\Psi_{t-1}(x_{t-1})}-1\right)^2$, and $ \ell^'_c(\beta^*_t) \triangleq  \nabla_{\beta_t}\ell_c(\beta_t)|_{\beta_t = \beta^*_t} $, if we set $\lambda_c = \frac{A}{\sqrt{n}}$, where $A$ is a positive constant, then $\sqrt{n}\ell^'_c(\beta^*_t) \xrightarrow{P} 0$.
\end{lemma}

\begin{proof}
\begin{equation*}
\begin{split}
     \sqrt{n_{t-1}}\ell^'_c(\beta_t) = 2A\left(\frac{\Psi_t(x_t)}{\Phi_t(x_t)\Psi_{t-1}(x_{t-1})}-1\right) \times \\ \left( \nabla_{\beta_t} \frac{\Psi_t(x_t)}{\Phi_t(x_t)\Psi_{t-1}(x_{t-1})}\right)  
\end{split}
\end{equation*}
By the law of large numbers,
\begin{equation*}
    \begin{split}
        &\frac{\Psi_t(x_t)}{\Phi_t(x_t)\Psi_{t-1}(x_{t-1})}\bigg|_{\beta_t=\beta_t^*} \xrightarrow{P} \\ 
        & \qquad \qquad \qquad 
        \frac{\mathbb{E}_{q_t}[\exp\{\psi_{\beta_t^*}(x)\}]}{\mathbb{E}_{q_t}[\exp\{\phi_{\beta^*_t}(x)\}]\mathbb{E}_{q_{t-1}}[\exp\{\psi_{{\beta}_{t-1}}(x)\}]} 
    \end{split}
\end{equation*}
Define $\tilde{r}_{t-1}(x) \triangleq \frac{\exp\{\psi_{\beta_{t-1}}(x)\}}{\mathbb{E}_{q_{t-1}}[\exp\{\psi_{\beta_{t-1}}(x)\}]}$, by the definition of $r^*_{s_t}(x)$ (\Cref{eq:def_optm}): 
\begin{equation*}
\begin{split}
\int \tilde{r}_{t-1}(x)  r^*_{s_t}(x) q_t(x) dx 
&= \int \tilde{r}_{t-1}(x) \frac{q_{t-1}(x)}{q_t(x)}  q_t(x) dx \\
& = \int \tilde{r}_{t-1}(x) q_{t-1}(x) dx = 1  
\end{split}
\end{equation*}
Substituting the right side of \Cref{eq:def_optm} into the left side of the above equation, we can get:
\begin{equation}
    \begin{split}
        & \frac{\mathbb{E}_{q_t}[\exp\{\psi_{\beta_t^*}(x)\}]}{\mathbb{E}_{q_t}[\exp\{\phi_{\beta^*_t}(x)\}]\mathbb{E}_{q_{t-1}}[\exp\{\psi_{{\beta}_{t-1}}(x)\}]} = 1 \\
    \end{split}
\end{equation}
then we prove this lemma. 
\end{proof}

\begin{customthm}{1}\label{thm1}
Suppose $\lambda_c = \frac{A}{\sqrt{n}}$, where $A$ is a positive constant, assume  $\ell^{''}_c(\beta^*_t) = o_p(1)$, $\hat{\beta_t}-\beta^*_t=o_p(1)$, $\mathcal{L}^{'''}_t(\tilde{\beta}_t) = O_p(1)$, where $\tilde{\beta}_t$ is a point between $\hat{\beta}_t$ and $\beta^*_t$, then $\sqrt{n} (\hat{\beta_t} - \beta_t^*) \leadsto \mathcal{N}(0, \nu^2)$, 
where 
\begin{equation}
    \begin{split}
      \nu^2 &=  I^{-1}_{\beta^*_t} + \mathbb{E}_{q_t}[\exp\{\phi_{\beta^*_t}(x)\}]^{-2} \times \\
     & \quad 
     I^{-1}_{\beta^*_t} Cov_{q_t}[\nabla_{\beta_t}\exp\{\phi_{\beta^*_t}(x)\}]
     I^{-1}_{\beta^*_t}
    \end{split}
\end{equation}
\end{customthm}
\begin{proof}
Applying Taylor expansion to \Cref{eq:def_hat} around $\beta_t^*$:
\begin{equation}
    \begin{split}
        &0 = \mathcal{L}^'_t(\hat{\beta}_t) =  \mathcal{L}^'_t({\beta}^*_t) + (\hat{\beta}_t - \beta^*_t)\mathcal{L}^{''}_t({\beta}^*_t) \\
        & \qquad \qquad \qquad + \frac{1}{2}(\hat{\beta}_t - \beta^*_t)^2 \mathcal{L}^{'''}_t({\tilde{\beta}_t}),\\
        &\sqrt{n}(\hat{\beta}_t - \beta^*_t) =  \frac{-\sqrt{n}\mathcal{L}^'_t({\beta}^*_t)}{\mathcal{L}^{''}_t({\beta}^*_t)+\frac{1}{2}(\hat{\beta}_t - \beta^*_t) \mathcal{L}^{'''}_t({\tilde{\beta}_t})}
    \end{split}
\end{equation}
where
\begin{equation}
    \mathcal{L}_t(\beta_t) = \ell_{r}(\beta_t) + \ell_{c}(\beta_t)
\end{equation}
As we assume $\ell^{''}_c(\beta^*_t) = o_p(1)$, according to \Cref{lemma2}, we get  $\mathcal{L}^{''}_t(\beta^*_t) \xrightarrow{P} -I_{\beta^*_t}$.  
Combine the results of \Cref{lemma1,lemma3} we prove the theorem.
\end{proof}

\begin{customcorl}{1}\label{corl1}
Suppose $p(x)$ and $\forall t, q_t(x)$ are from the exponential family, define $r^*_{s_t}(x) = \exp\{\phi_{\beta^*_t}(x)\}$,  $\phi_{\beta^*_t}(x) = \beta^*_{t} T(x)+C$,  $T(x)$ is a sufficient statistic of $x$, $C$ is a constant, then  $\sqrt{n} (\hat{\beta_t} - \beta_t^*) \leadsto \mathcal{N}(0, \nu_e^2)$, where $T(x)$ is a column vector, $T(x)^2 = T(x)T(x)^T$, $I_{\beta^*_t}=Cov_{q_{t-1}}[{T}(x)]$:
\begin{equation}\label{eq:corl1}
    \begin{split}
        \nu_e^2 &= I^{-1}_{\beta^*_t} + I^{-1}_{\beta^*_t}  ({\mathbb{E}_{q_{t-1}}[r^*_{s_t}(x){T}(x)^2]-\mathbb{E}_{q_{t-1}}[{T}(x)]^2})I^{-1}_{\beta^*_t}
    \end{split}
\end{equation}
\end{customcorl}

\begin{proof}
 Because $\phi^*_{t}(x) \triangleq \psi_{\beta_t^*}(x) - \psi_{{\beta}^*_{t-1}}(x)$, then $\nabla_{\beta_t}\phi_{\beta_t}^*(x) = T(x)$, we have     
\begin{equation}
    \begin{split}
    & I_{\beta^*_t} =  Cov_{q_{t-1}}[\nabla_{\beta_t}\phi_{\beta_t}^*(x)] = Cov_{q_{t-1}}[T(x)], \\
    \end{split}
\end{equation}
Because $r^*_{s_t}(x) = \exp\{\phi^*_{t}(x)\}$, 
\begin{equation*}
 \mathbb{E}_{q_t}[\exp\{\phi_{\beta_t}^*(x)\}] = \mathbb{E}_{q_t}[r^*_{s_t}(x)] = 1, 
\end{equation*}
In addition,
\begin{equation}
    \begin{split}
     &Cov_{q_t}[\nabla_{\beta_t}\exp\{\phi_{\beta_t}^*(x)\}] = Cov_{q_t}[r^*_{s_t}(x) T(x)] \\ 
    & \qquad = 
    \mathbb{E}_{q_t}[(r^*_{s_t}(x) T(x))^2] - \mathbb{E}_{q_t}[r^*_{s_t}(x) T(x)]^2    
    \end{split}
\end{equation}
where 
\begin{equation}
    \begin{split}
    & \mathbb{E}_{q_t}[(r^*_{s_t}(x) T(x))^2] = \int q_t(x) (r^*_{s_t}(x) T(x))^2 dx \\
    & = \int q_{t-1}(x)r^*_{s_t}(x) T(x)^2 dx = \mathbb{E}_{q_{t-1}}[r^*_{s_t}(x) T(x)^2], \\
    &\mathbb{E}_{q_t}[r^*_{s_t}(x) T(x)] = \int q_t(x) r^*_{s_t}(x) T(x) dx \\
    &= \int q_{t-1}(x) T(x) dx = \mathbb{E}_{q_{t-1}}[T(x)]
    \end{split}
\end{equation}
Substitute above results into Theorem 1, we prove the corollary.  
\end{proof}

\section{Experimental settings}\label{sec:experiment_setting}

We elaborate the experimental settings of the ratio estimators, feature generators, and \acsp{GAN} in our experiments here. The source code will be made public upon acceptance. 

\subsection{Configuration of ratio estimators}

The ratio estimator we used in all experiments is the log-linear model as
defined in Eq. 9 of Sec. 3, and $\psi(\cdot)$ is a neural network
with two dense layers, each having 256 hidden units. It is trained by Adam
optimizer, and batch size is 2000, learning rate is $10^{-5}$. $\lambda_c$ is $10$ and $100k$ for the single and multiple original distributions,  respectively, where $k$ is the number of joined original distributions. 

\subsection{Configuration of feature generators}
The classifier used to extract features on both datasets has two convolutional
layers with filter shape [4,4,1,64], [4,4,64,128] respectively, strides are all
[1,2,2,1], and two dense layers
with 6272 and 64 hidden units respectively.  Batch normalization is performed on the second convolutional layer and the first dense layer.
The encoder of the \acs{CVAE} has two dense layers with 512 and 256 hidden units respectively, output
dimension is 64. The decoder has two dense layers each with 256 hidden units,
output dimension is 784. Activation function of all hidden layers is ReLU. Both feature generators are trained with L2 regularization.


\subsection{Configuration of GANs}

All GANs are trained with a discriminator having the same architecture as the
classifier described above, except the last dense layer has 1024 hidden unites;
a generator having two convolutional layers with filter shape
[4,4,64,128] and [4,4,1,64], respectively, two dense layers with 1024 and 6272 hidden units
respectively, batch normalization is applied to two dense layers. Activation
function is leaky ReLu \citep{xu2015empirical} for all hidden layers. All \acp{GAN} are trained by Adam optimizer, batch size is 64, learning rate is 0.0002 and 0.001 for the 
discriminator and generator respectively. 
The random input of the generators has 64 dimensions for both MNIST and Fashion-MNIST tasks.

\section{Evaluating generative models using \fdiv{}}\label{sec:general_toy_eval}

We compare \fdiv{} with \acs{FID} \citep{heusel2017gans}, \acs{KID} \citep{binkowski2018demystifying} and \acs{PRD} \citep{sajjadi2018assessing} in several experiments with toy data and a high-dimensional dataset of the real world. Through these experiments, we show that \fdiv{} can be effective measures of generative models and one may obtain richer criteria by them.

\subsection{Demo experiments with MNIST}
We present demo experiments to show the differences between \fdiv{} and \acs{FID}, \acs{KID} in terms of evaluating generative models. We demonstrate the experiment results through two most popular members of \fdiv{}: KL-divergence and reverse KL-divergence. 

\begin{figure}[t!]
\begin{subfigure}{0.425\linewidth}
\includegraphics[width=\linewidth,trim={0.cm 0.cm 0.cm 0.cm},clip]{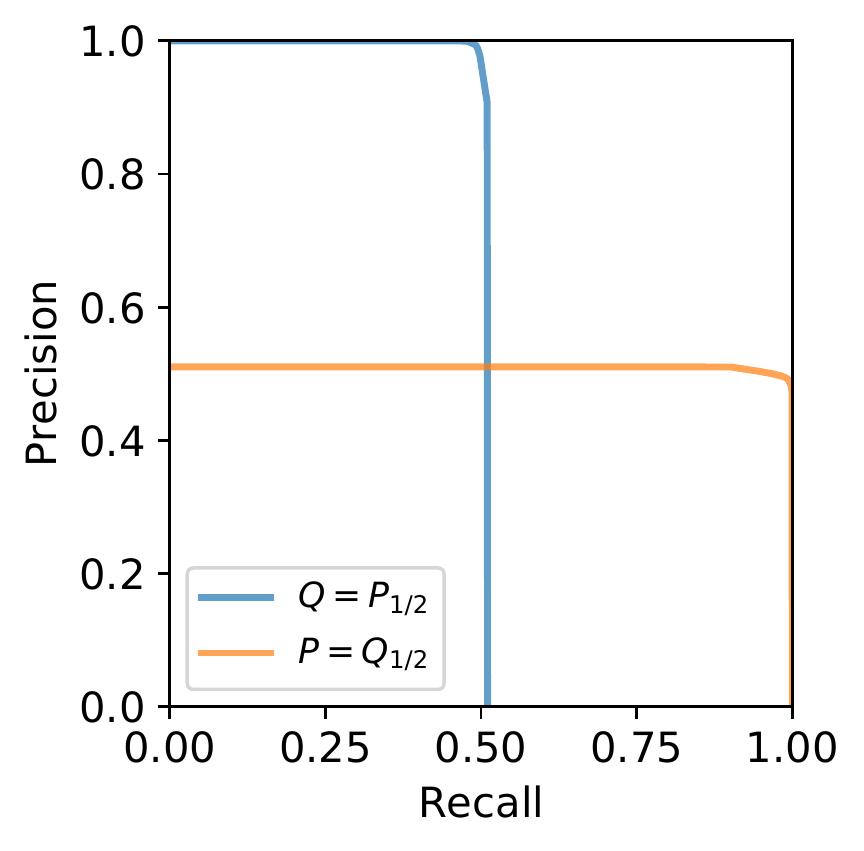}
\caption{PRD of the demo experiment in \Cref{tab:toy_mnist}}
\label{fig:toy_prd}
\end{subfigure}
\hfill
\begin{subfigure}{0.5\linewidth}
    \includegraphics[width=\linewidth,trim={1.cm 0.cm 0.cm 0.cm},clip]{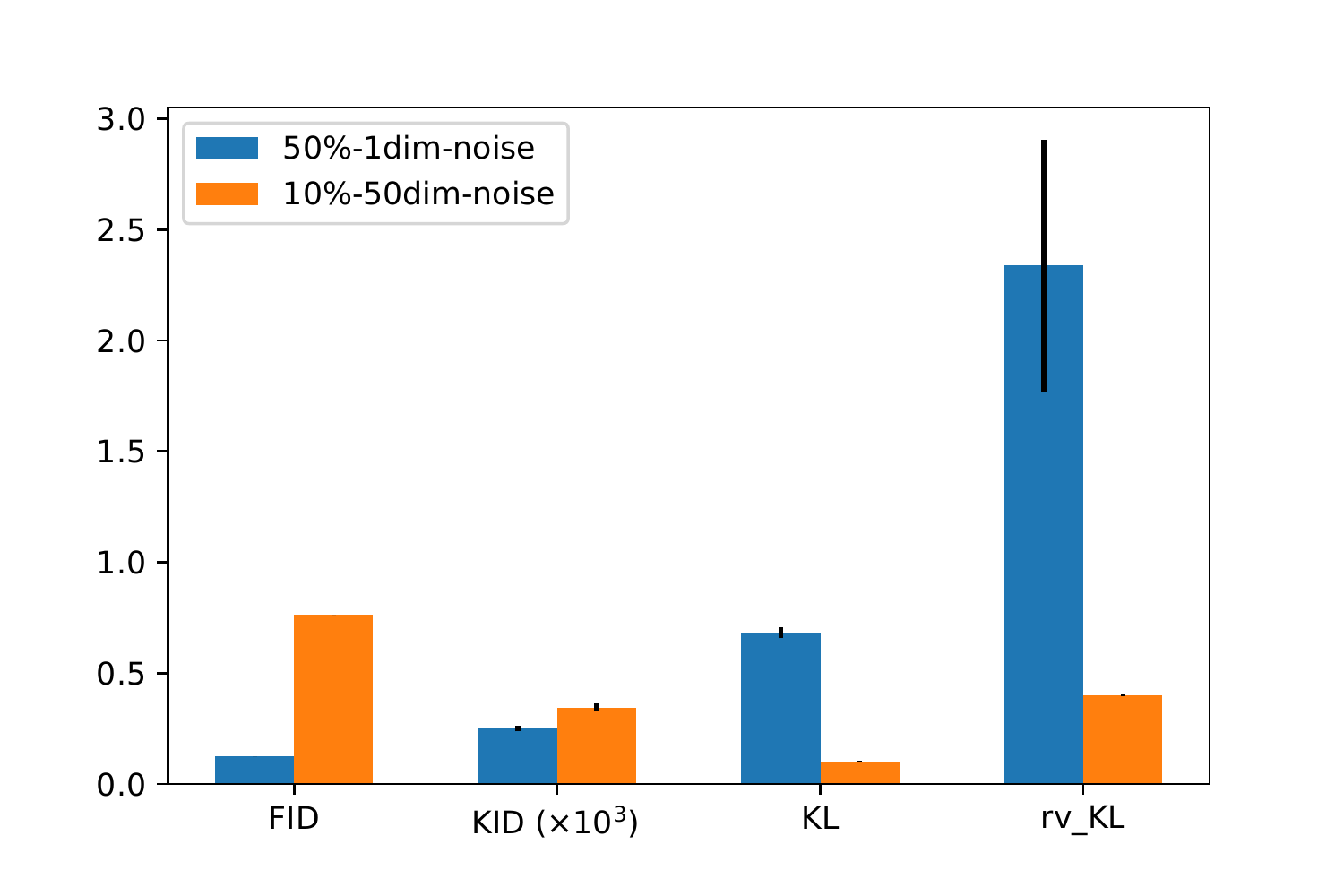} 
    \caption{Demo experiments with different types of noise injected into MNIST (error bars from 5 runs).}
    \label{fig:toy_noise}
\end{subfigure}
\caption{Experimental results of demo experiments}
\end{figure}

In the first experiment, We have two cases: (i) the compared distribution $P$ contains half of the classes of MNIST, and the evaluated distribution $Q$ includes all classes of MNIST;
 (ii) the reverse of (i). We obtain density ratios by \acs{KLIEP} and then estimate \fdiv{} by the ratios. The results are shown in
 \Cref{tab:toy_mnist}, with \acs{PRD} curves displayed in \Cref{fig:toy_prd}. 
 Since the objective of \acs{KLIEP} is not symmetric, the estimated KL divergences are not symmetric when switching the two sets of samples. 
 As we can see, $D_{KL}(P||Q)$ prefers $Q$ with larger recall and vice versa. Neither
 FID nor KID are able to discriminate between these two scenarios. 

In the second experiment, we show that \fdiv{} may provide different opinions with \acs{FID} and \acs{KID} in certain circumstances because \acs{FID} and \acs{KID} are based on \acs{IPM} \citep{sriperumbudur2012empirical} which focus on parts with the  most probability mass whereas \fdiv{} are based on density ratios which may give more attention on parts with less probability mass (due to the ratio of two small values can be very large). To show this, we simulate two sets of noisy samples by injecting two different types of noise into MNIST data and evaluate them as the model samples on the pixel feature space (which is 784 dimensions). Regarding the first type of noise, we randomly choose 50\% samples and 1 dimension to be corrupted (set the pixel value to 0.5); for the second one, we randomly choose 10\% samples and 50 dimensions to be corrupted. The results are shown in \Cref{fig:toy_noise}, in which KL-divergence and reverse KL-divergence disagree with \acs{FID} and \acs{KID} regarding which set of samples is better than the other.
\setlength{\tabcolsep}{3pt}
\begin{table}[h]
    \centering
    \small
    \captionof{table}{ Results of the first demo experiment. $P = Q_{1/2}$ implies case (i), $Q = P_{1/2}$ implies case (ii). Standard deviations are from 5 runs.}   
    \begin{tabular}{ccccc}
    \toprule
    & FID                  & KID                 & $D_{KL}(P||Q)$    
    & $D_{KL}(Q||P)$      \\
    \midrule
    $P = Q_{1/2}$ & 50.39 $\pm$ 0.00 & 2.04 $\pm$ 0.01 & 0.67 $\pm$ 0.00 & 3.78  $\pm$ 1.22  \\
    $Q = P_{1/2}$ & 50.39 $\pm$ 0.00 & 2.03 $\pm$ 0.02 & 2.49 $\pm$ 0.30 & 2.38 $\pm$ 1.78 \\
    \bottomrule
    \end{tabular}
    \label{tab:toy_mnist}
\end{table}

\subsection{Evaluating on   high-dimensional data by \fdiv{}} \label{sec:ffhq}

In order to show that \fdiv{} can work with high dimensional data as the same as \acs{FID} and \acs{KID}, we also conducted an experiment with samples of StyleGAN\footnote{\url{https://github.com/NVlabs/stylegan}} trained by FFHQ dataset \citep{karras2019style}. We estimate \fdiv{} on the inception feature space with 2048 dimensions. The sample size is 50000 and we compare the model samples with real samples. We see that the \fdiv{} give reasonable results with small variance (\Cref{tab:stylegan}),   indicating it is capable of evaluating generative models with high dimensional data. 

\begin{table}[h]
\small
\centering
\caption{Evaluating StyleGAN on FFHQ dataset using \fdiv{}}
\label{tab:stylegan}
\begin{tabular}{lll}
\toprule
          & StyleGAN        & Real samples      \\
\midrule
KL        & $2.47 \pm 0.02$ & $0.02 \pm 9.1e-4$ \\
rv\_KL    & $3.29 \pm 0.18$ & $0.02 \pm 9.3e-4$ \\
JS        & $0.86 \pm 0.01$ & $0.01 \pm 4.5e-4$ \\
Hellinger & $1.04 \pm 0.02$ & $0.01 \pm 4.6e-4$ \\
\bottomrule
\end{tabular}
\end{table}

\section{Experiment results on MNIST}\label{sec:mnist}

\begin{figure}[t!]
    \centering
    \begin{subfigure}{0.425\linewidth}
    \includegraphics[width=\linewidth,trim={1.7cm 1.1cm 2.2cm 1.7cm},clip]{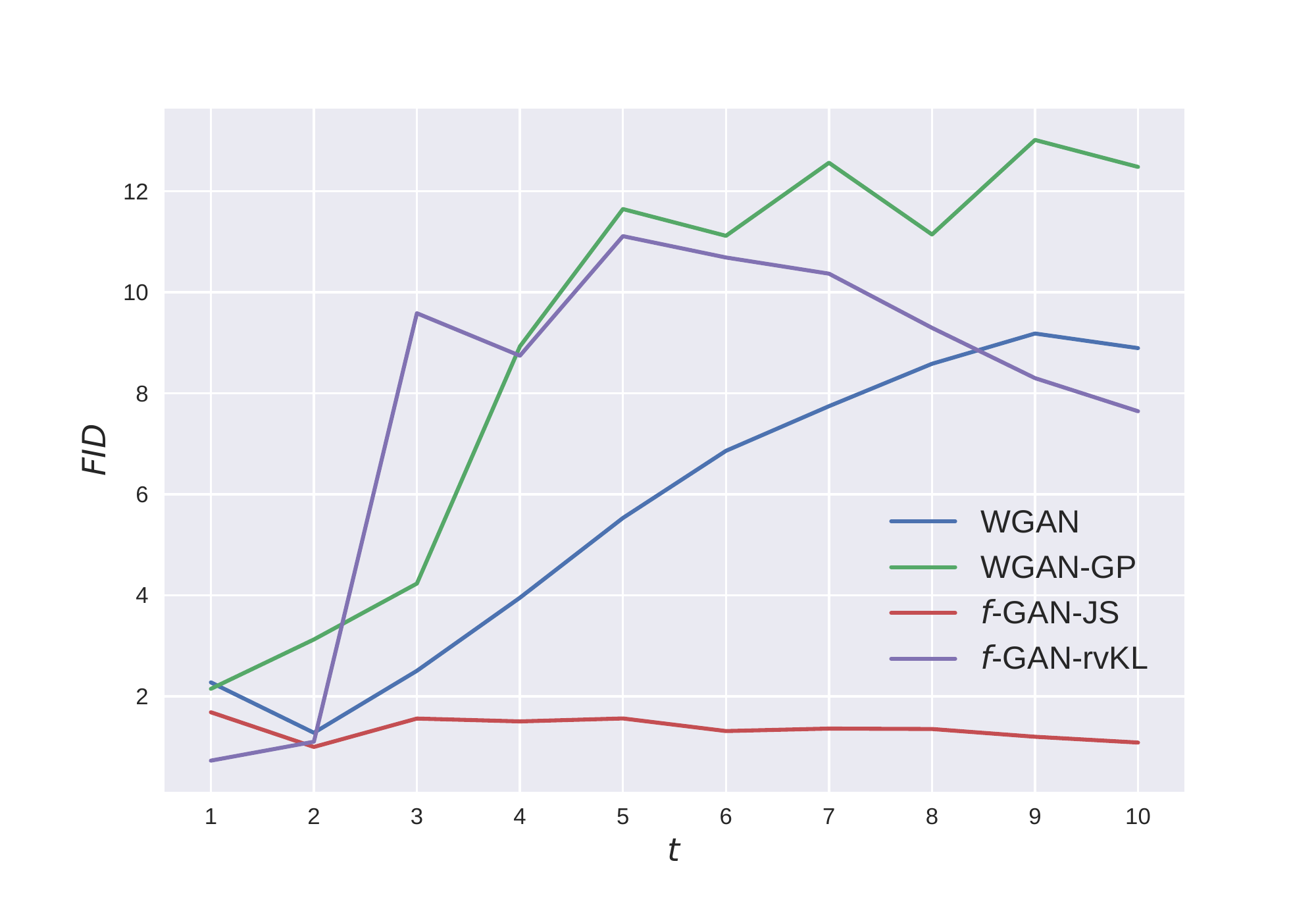}
    \caption{FID}
    \label{fig:cl_fid}
    \end{subfigure} 
    \begin{subfigure}{0.425\linewidth}
    \includegraphics[width=\linewidth,trim={1.7cm 1.1cm 2.2cm 1.7cm},clip]{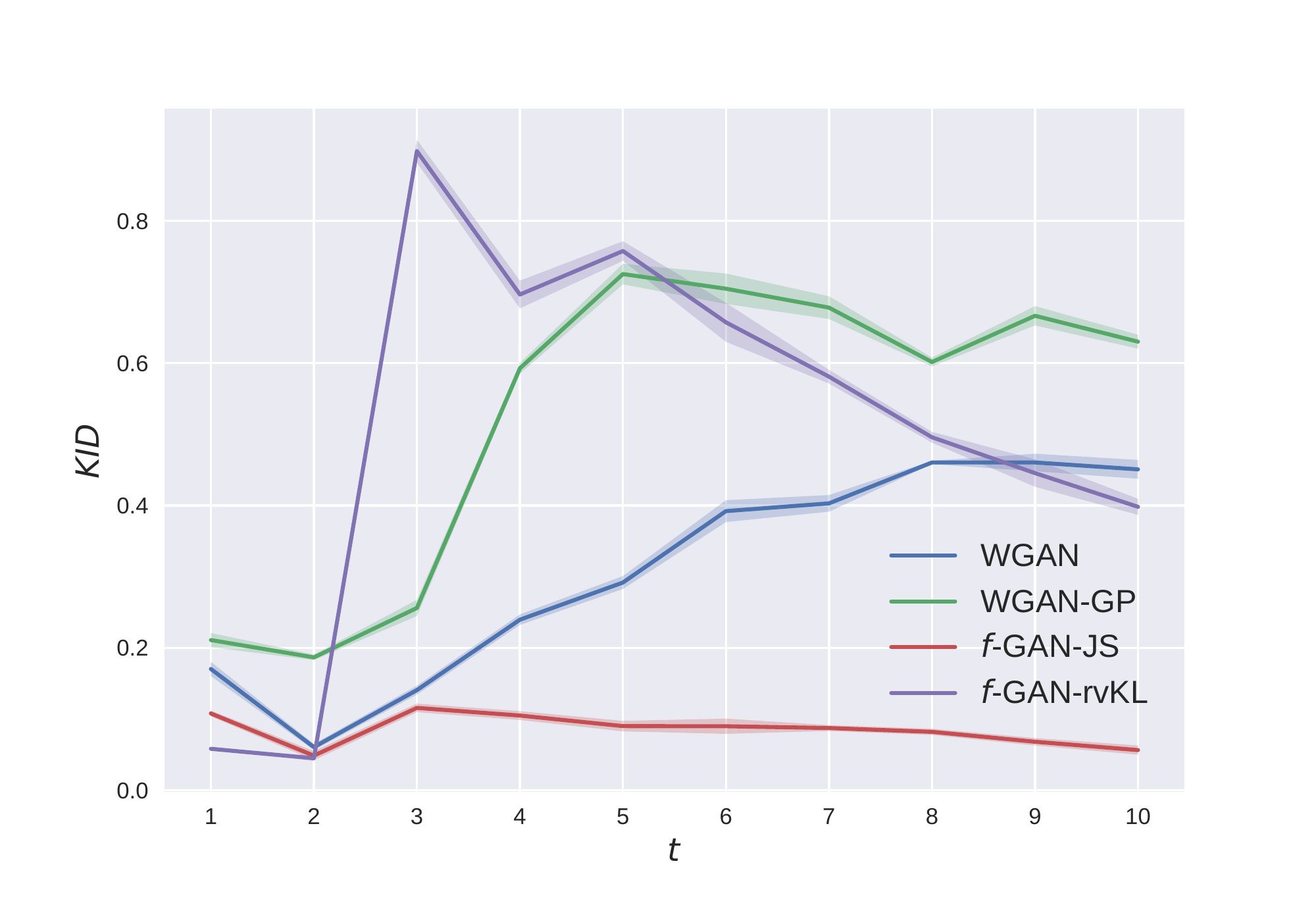}
    \caption{KID}
    \label{fig:cl_kid}
    \end{subfigure}
    \\%
    \begin{subfigure}{0.425\linewidth}
    \includegraphics[width=\linewidth,trim={1.4cm 1.cm 1.8cm 1.2cm},clip]{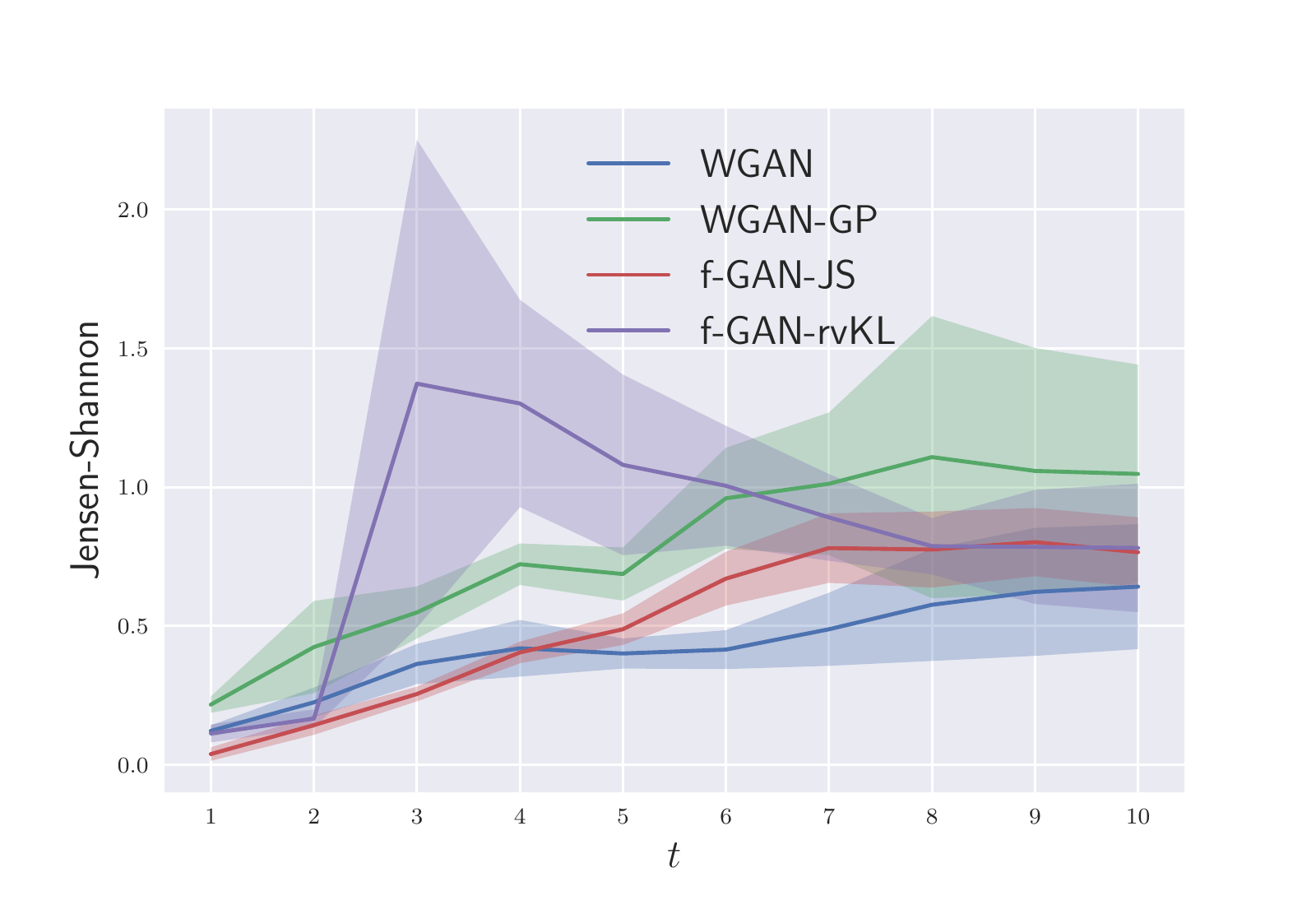}
    \caption{KLpq}
    \label{fig:mnist_klpq}
    \end{subfigure}
    \begin{subfigure}{0.425\linewidth}
        \includegraphics[width=\linewidth,trim={1.4cm 1.cm 1.8cm 1.2cm},clip]{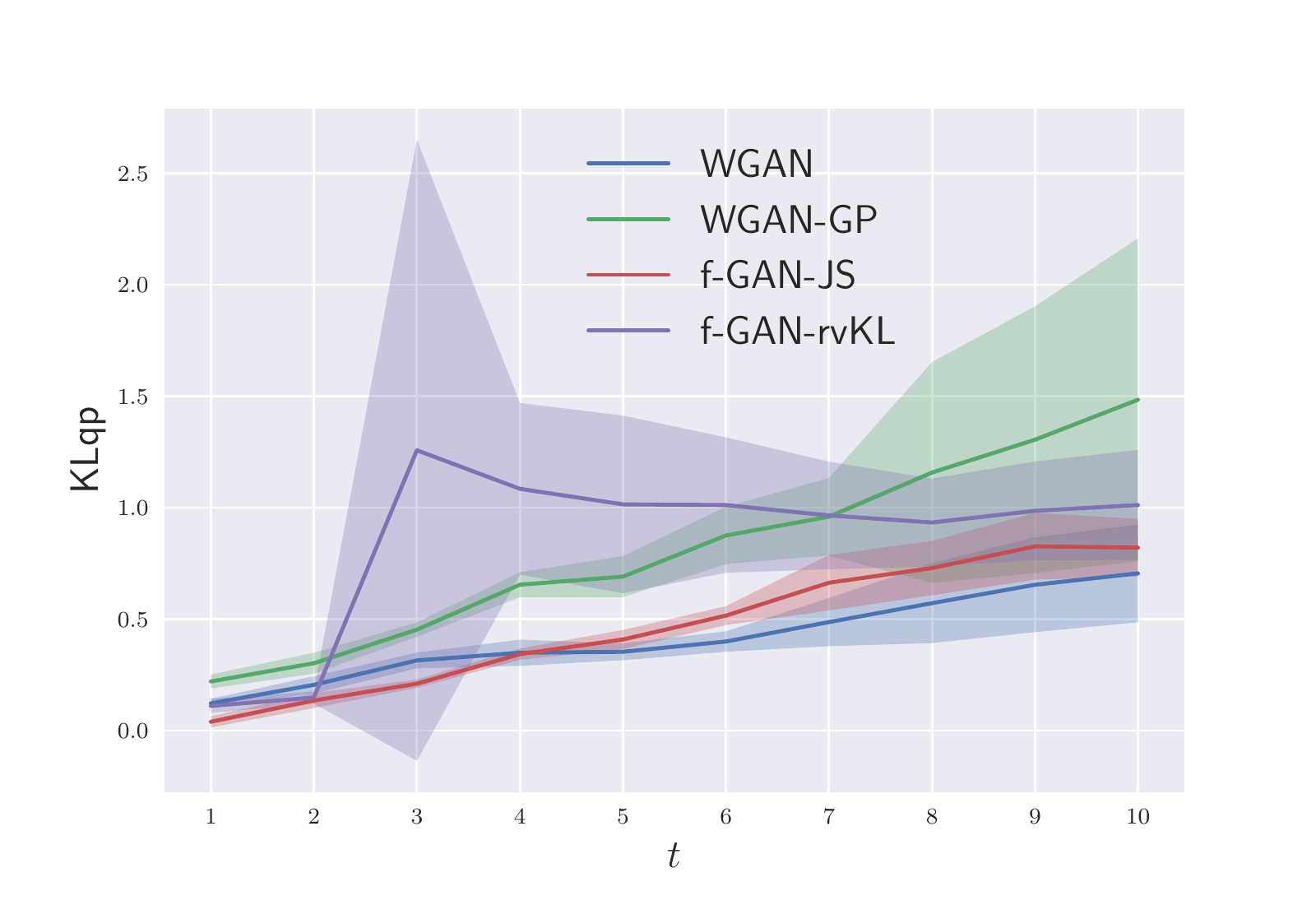}
        \caption{KLqp}
        \label{fig:mnist_klqp}
        \end{subfigure}
        \\%
    \begin{subfigure}{0.425\linewidth}
    \includegraphics[width=\linewidth,trim={1.4cm 1.cm 1.8cm 1.2cm},clip]{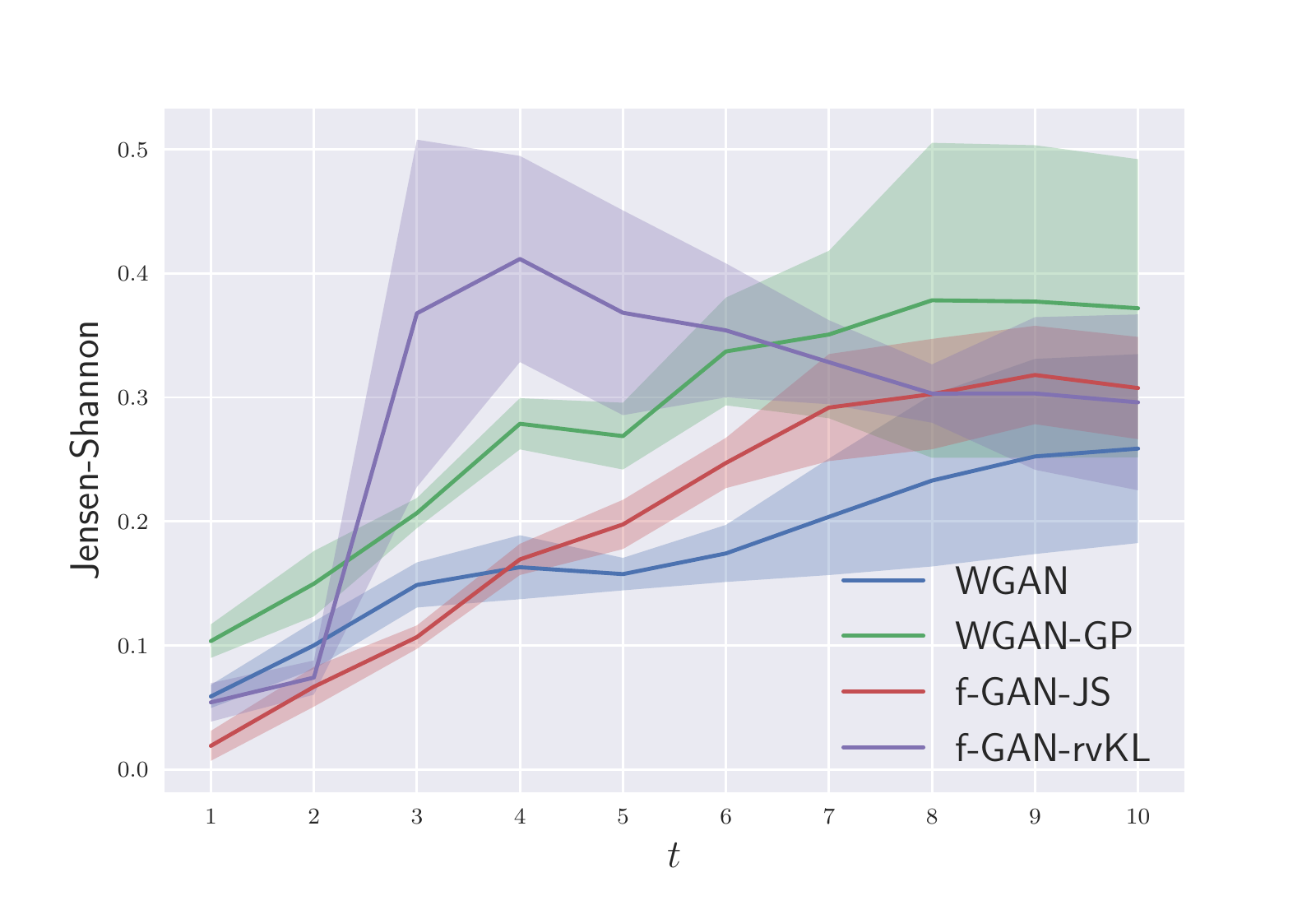}
    \caption{JS}
    \label{fig:mnist_js}
    \end{subfigure}
    \begin{subfigure}{0.425\linewidth}
        \includegraphics[width=\linewidth,trim={1.4cm 1.cm 1.8cm 1.2cm},clip]{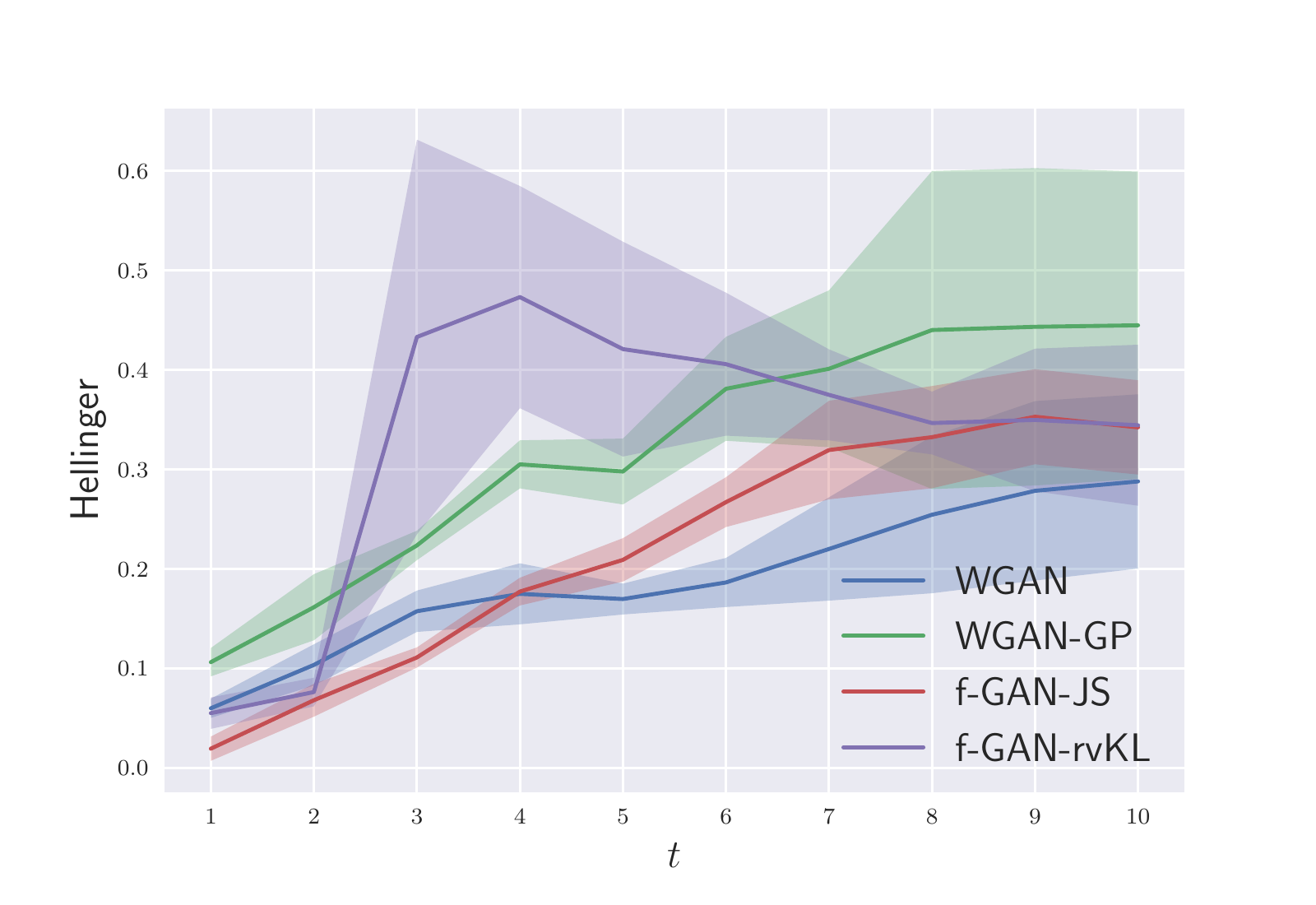}
        \caption{Hellinger}
        \label{fig:mnist_hg}
        \end{subfigure}
\caption{Evaluating \acp{GAN} in \cl{} on MNIST, features for FID
and KID are extracted from the classifier,
features for \fdiv{} are generated by CVAE. The shaded area
are plotted by standard deviation of 10 runs.}
\label{fig:mnist_cl}
\end{figure}

\begin{figure}[!t]
    \centering
    \begin{subfigure}{0.48\linewidth}
    \includegraphics[width=\linewidth,trim={2.cm 2.cm 2.cm 2.cm},clip]{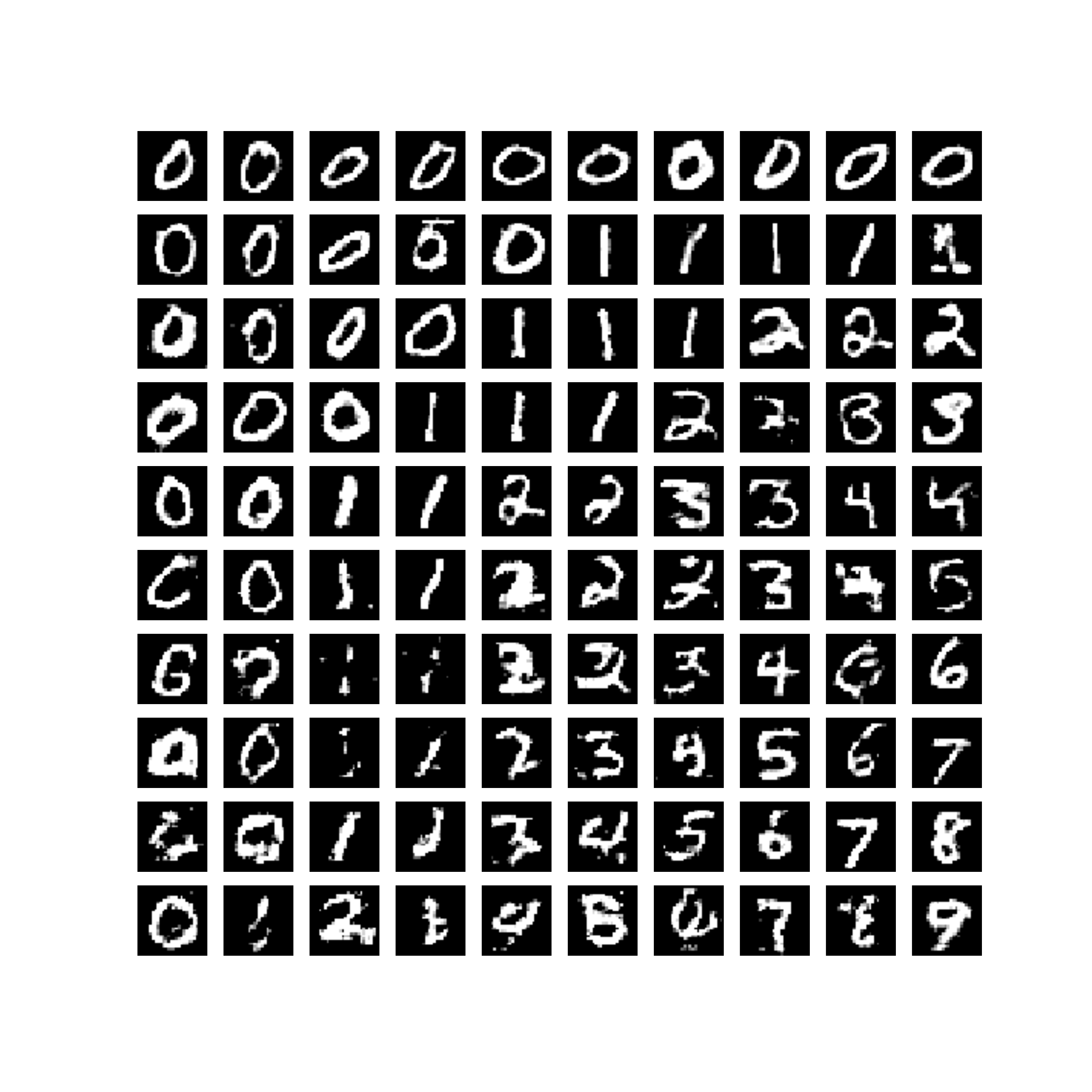}
    \caption{WGAN}
    \label{fig:wgan_mnist}
    \end{subfigure}
    \begin{subfigure}{0.48\linewidth}
    \includegraphics[width=\linewidth,trim={2.cm 2.cm 2.cm 2.cm},clip]{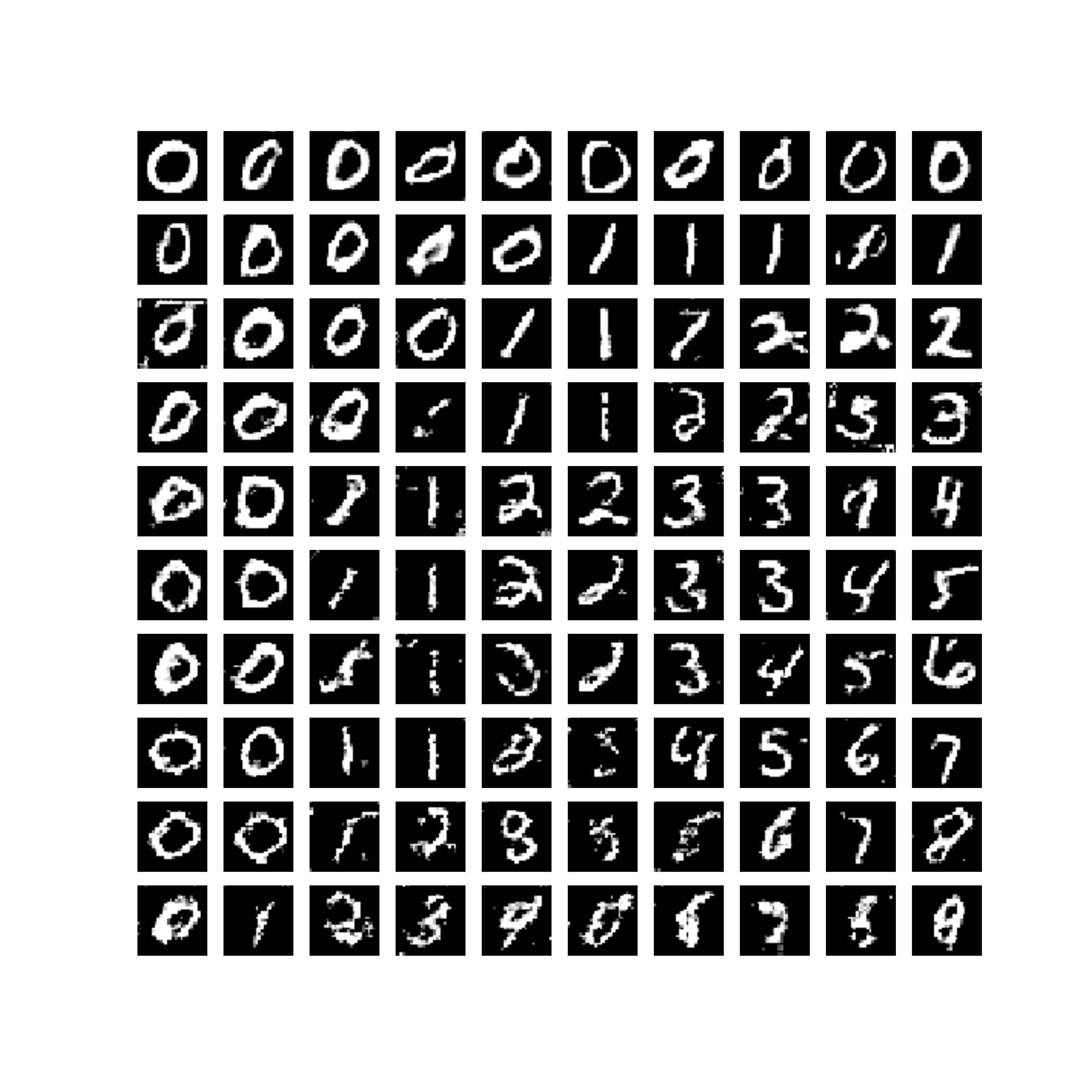}
    \caption{WGAN-GP}
    \label{fig:wgangp_mnist}
    \end{subfigure}
    \\%
    \begin{subfigure}{0.48\linewidth}
    \includegraphics[width=\linewidth,trim={2.cm 2.cm 2.cm 2.cm},clip]{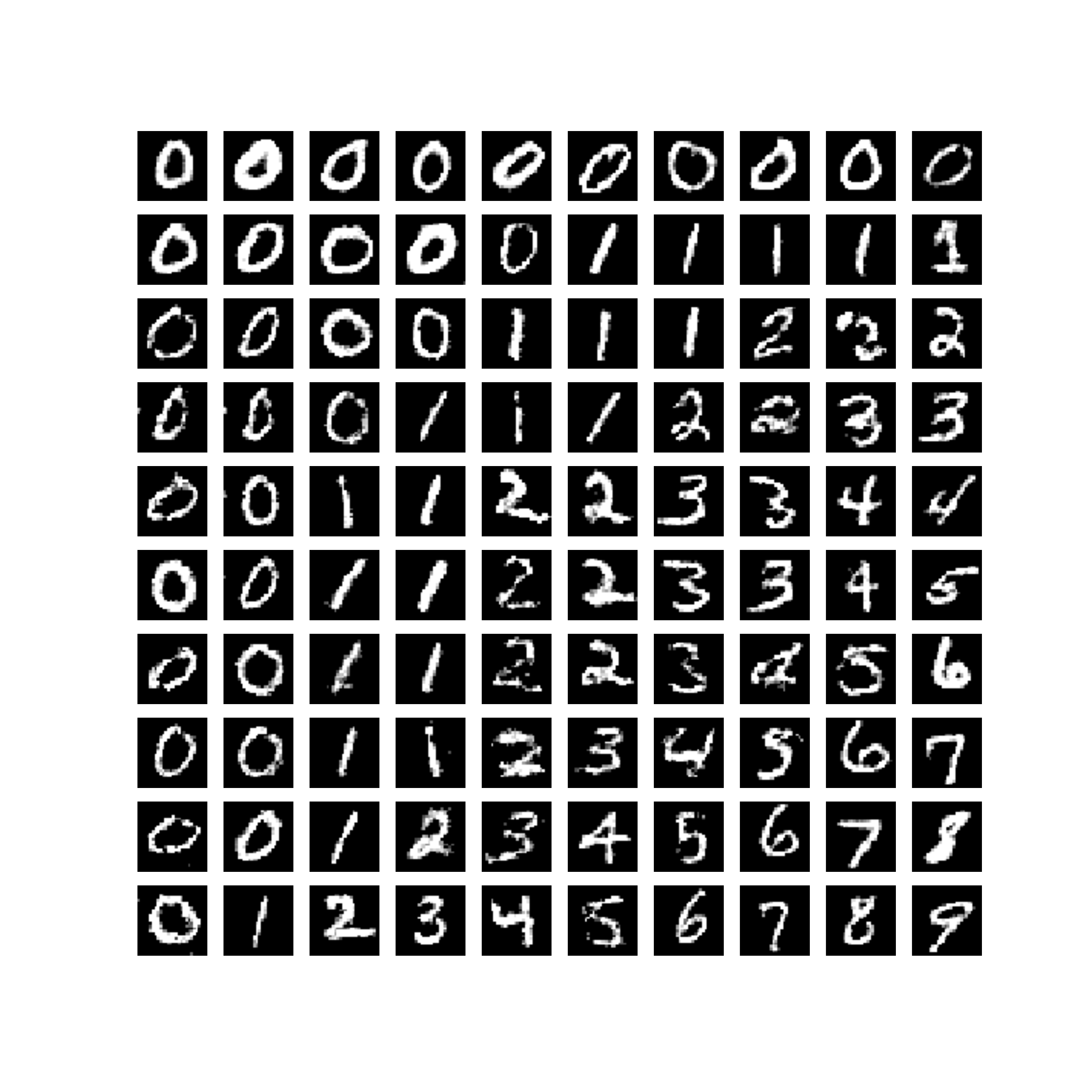}
    \caption{\fgan-JS}
    \label{fig:js_mnist}
    \end{subfigure}
    \begin{subfigure}{0.48\linewidth}
    \includegraphics[width=\linewidth,trim={2.cm 2.cm 2.cm 2.cm},clip]{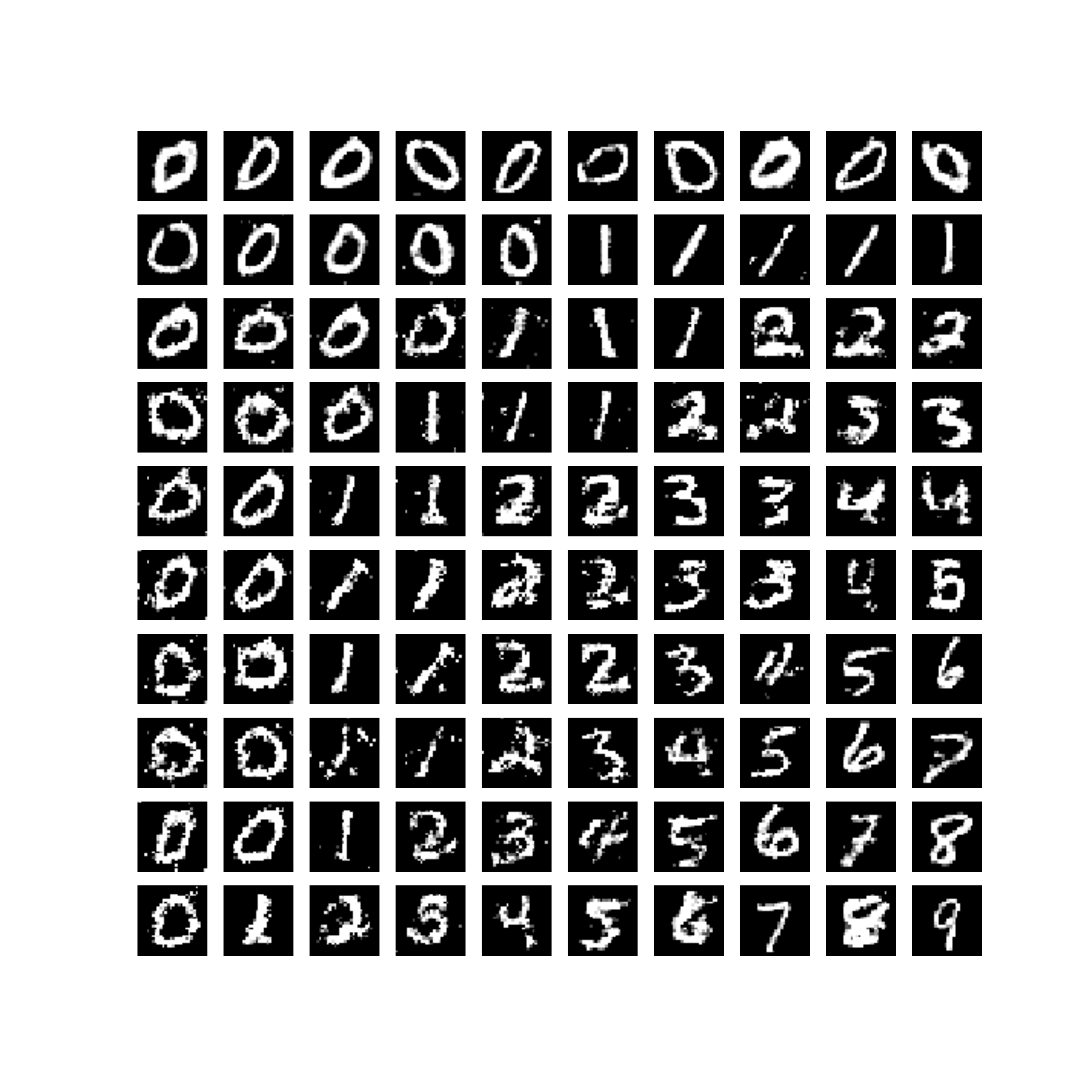}
    \caption{\fgan-rvKL}
    \label{fig:rvkl_mnist}
    \end{subfigure}
\caption{MNIST samples generated by evaluated GANs in \cl{}. In each figure, each row displays figures generated at each task,the order is from the top to bottom.}
\label{fig:mnist_task_samples}
\end{figure}

We show experiment results on MNIST in \Cref{fig:mnist_cl,fig:mnist_task_samples}, the settings of these experiments are the same as experiments with Fashion-MNIST.


